\documentclass{article}
\PassOptionsToPackage{dvipsnames}{xcolor}
\usepackage{arxiv}

\usepackage{fontspec}

\defaultfontfeatures{Ligatures=TeX}

\usepackage{amsmath}
\usepackage{amssymb}
\usepackage{amsfonts}
\usepackage{bm}
\usepackage{mathtools}
\usepackage{nicefrac}

\usepackage{graphicx}

\graphicspath{{./fig/}}

\usepackage{tikz}
\usetikzlibrary{
  arrows.meta,
  positioning,
  shapes.geometric,
  fit,
  backgrounds,
  calc
}

\usepackage{float}

\usepackage{booktabs}
\usepackage{multirow}
\usepackage{array}
\usepackage{tabularx}
\usepackage{makecell}

\newcolumntype{C}{>{\centering\arraybackslash}X}

\newcolumntype{L}[1]{%
  >{\raggedright\arraybackslash}p{#1}%
}

\newcolumntype{P}[1]{%
  >{\centering\arraybackslash}p{#1}%
}

\usepackage{caption}
\usepackage{subcaption}

\usepackage[
  ruled,
  vlined,
  linesnumbered
]{algorithm2e}

\usepackage{enumitem}

\usepackage{soul}
\sethlcolor{yellow}

\usepackage{microtype}

\usepackage{url}
\usepackage{hyperref}

\hypersetup{
  colorlinks=true,
  linkcolor=NavyBlue,
  citecolor=BrickRed,
  urlcolor=NavyBlue
}

\newcommand{\method}{\textsc{LG-VLN}}

\title{LG-VLN: A Zero-Shot Vision-and-Language Navigation Framework with LangGraph State Orchestration}

\author{
  Jianhe Zhao$^{1}$,
  Yanhua Qiu$^{1}$\thanks{Corresponding author: Yanhua Qiu. E-mail: 2025202140028@whu.edu.cn}, 
  Zhiyu Zhang$^{2}$,
  Zibo Zhao$^{3}$, 
  Jinhua Xie$^{1}$\\
  $^{1}$School of Geodesy and Geomatics, Wuhan University, Wuhan, China \\
  $^{2}$School of Computer Science and Technology, Huazhong University of Science and Technology, Wuhan, China \\
  $^{3}$Academy of Advanced Interdisciplinary Studies, Wuhan University, Wuhan, China
}

\date{}

\begin{document}
\maketitle

\begin{abstract}
Continuous-environment vision-and-language navigation (VLN-CE) requires interpreting natural-language instructions in unseen 3D environments and executing continuous low-level actions. Existing methods often depend on LiDAR, panoramic cameras, or extra sensors; separate geometric-mapping and semantic-navigation visual representations can cause long-trajectory spatial-semantic inconsistencies. We propose LG-VLN, a monocular zero-shot framework with shared visual features and LangGraph-based state orchestration. An online feed-forward 3D reconstruction network predicts depth, camera poses, and dense point clouds for agent-pose estimation and global map fusion. Geometry and navigation share dense CleanDIFT features: semantic consistency rejects incorrect inter-frame correspondences, while target-instance constraints define visual references whose similarity combines with local BLIP-2 image-text relevance to form a semantic value map. LangGraph represents instruction parsing, geometric perception, semantic value updates, path planning, action execution, and failure recovery as a directed state graph with conditional transitions, persistent state, and modular recovery mechanisms. On a fixed 550-episode subset of the R2R-CE val-unseen split, LG-VLN achieves 21.3\% success and 12.1\% success weighted by path length. Ablations show shared semantic features improve navigation, further boosted by combining visual similarity and image-text relevance. Results establish shared visual representations and explicit state orchestration as effective for zero-shot VLN-CE using monocular RGB alone. Code will be publicly released for reproducibility.
\end{abstract}


\section{Introduction}
\label{sec:intro}

Vision-and-Language Navigation (VLN) requires an agent to interpret natural-language instructions and reach a target location in an unfamiliar three-dimensional environment, making it an important task in embodied intelligence \cite{anderson2018vln}. Early VLN methods typically operate on discrete navigation graphs, where agents move between predefined waypoints along known topological connections \cite{anderson2018vln}. Vision-and-Language Navigation in Continuous Environments (VLN-CE), in contrast, assumes neither a prior map nor predefined navigation points. Agents must instead generate low-level control actions directly from sensory observations while navigating through continuous space \cite{krantz2020vlnce,savva2019habitat}. This setting more closely reflects the conditions faced by real-world robots and requires close integration of environmental perception, instruction understanding, and action selection \cite{krantz2020vlnce}. Recent advances in vision-language foundation models have also increased interest in zero-shot VLN-CE \cite{zhou2023navgpt,yokoyama2024vlfm,long2024instructnav,qiao2024opennav,chen2025canav}. These methods use pretrained models to parse instructions, identify semantically relevant locations, and select actions without access to prior maps or task-specific training, allowing them to navigate unseen environments and follow previously unseen instructions.

Despite recent progress, existing zero-shot VLN-CE methods still face several limitations in system design. Many approaches rely on additional sensors or observation modalities, such as depth sensors, panoramic images, and odometry, to recover environmental geometry, generate candidate waypoints, and estimate the robot's pose \cite{krantz2020vlnce,an2024etpnav,yokoyama2024vlfm,chen2025canav}. While these inputs can improve spatial perception and path planning, they also increase system complexity and the cost of real-world deployment. Another limitation lies in how semantic value maps are constructed. Most existing methods measure the similarity between the language instruction and either the current observation or local image regions, then project the resulting scores onto a two-dimensional map. Directly mapping language similarity to 2D regions often produces coarse spatial representations in which neighboring pixels receive nearly identical values. This limits fine-grained semantic grounding and makes it difficult to distinguish the target from contextual or irrelevant regions within the same view \cite{yokoyama2024vlfm,chen2025canav,chen2025aoplanner}. Some methods aggregate historical observations through weighted fusion across frames, but they generally do not maintain explicit correspondences between pixels or target regions over time, nor do they explicitly enforce temporal consistency \cite{chen2025canav,chen2025aoplanner}. Changes in viewpoint, target scale, or occlusion can therefore cause semantic values to drift, making them less reliable for fine-grained local action selection. These local perception issues are compounded by limited system-level control over long-horizon navigation. VLN-CE tasks may involve multiple sub-instructions that must be completed in sequence, while errors such as false target detections, unreachable subgoals, local planning failures, and execution failures can occur along the way \cite{song2025lhvln,wei2025streamvln}. Previous studies have addressed parts of this problem through instruction decomposition, progress estimation, historical memory, phase transitions, and replanning \cite{zhan2024mcgpt,zeng2026janusvln,wei2025streamvln}. However, without a unified and verifiable state representation, it remains difficult for the system to identify persistent errors and recover from false detections, unreachable subgoals, or repeated planning failures. As instructions become longer and navigation episodes extend over greater distances and durations, errors in perception, planning, and control can accumulate, reducing the stability of the overall system \cite{song2025lhvln}.

Beyond perception and planning, a VLN-CE system must also track sub-instruction progress, map states, and action feedback throughout navigation, while recovering from failures when necessary. These elements can be maintained in a shared state that records task progress, the active sub-instruction, the current map state, and action history. With this representation, perception, mapping, constraint evaluation, path planning, and action execution can be coordinated through well-defined state transitions. Recent studies have distributed navigation functions across multiple collaborating agents or exposed them through tool-calling interfaces, suggesting that modular workflows provide a practical structure for embodied navigation systems \cite{macnav2026,agenticnav2026}.

We implement this stateful navigation workflow using LangGraph \cite{langgraph}. Shared state, conditional transitions, controlled loops, and checkpointing allow the system to explicitly track sub-instruction transitions, perception scheduling, navigation phases, and recovery states \cite{langgraph,agenticai2026}. We define the node functions, shared state, and transition conditions according to the requirements of VLN-CE and organize them into a unified workflow. LangGraph primarily serves to coordinate state management across modules and to maintain an explicit record of module execution and state changes during navigation.

In summary, we propose LG-VLN, a monocular zero-shot Vision-and-Language Navigation framework based on shared dense visual features. Unlike existing methods that use separate feature extraction pipelines for 3D reconstruction and semantic navigation, LG-VLN supports both tasks with a common set of dense diffusion features. During mapping, semantic consistency is used to reject incorrect cross-frame correspondences, reducing point-cloud outliers and improving global geometric consistency. During navigation, an open-vocabulary vision-language model identifies semantic reference pixels according to the active sub-instruction, and the diffusion features propagate this information across frames to construct a dense semantic value map. At the system level, LG-VLN organizes perception, constraint evaluation, planning, and execution as computational nodes within a shared-state workflow controlled by a graph-structured state machine. This architecture coordinates sub-instruction transitions, supports failure recovery, and provides a unified mechanism for managing state throughout navigation.

The main contributions are summarized as follows:

1. LG-VLN operates using only monocular RGB observations, without requiring active depth sensors, LiDAR, or panoramic cameras. The depth estimates, camera poses, and dense point clouds needed for navigation are produced online by a feed-forward three-dimensional reconstruction network, reducing hardware requirements and calibration complexity.

2. LG-VLN also uses a shared visual representation for geometric reconstruction and semantic navigation. The same dense diffusion features are used to filter incorrect correspondences during mapping and to construct the semantic value map during navigation. Sharing the feature space allows geometric reasoning and semantic guidance to remain consistent across these two stages.

3. For long-horizon navigation, we formulate VLN-CE as a stateful workflow and implement its control logic as a graph-structured state machine using LangGraph. The workflow explicitly tracks sub-instruction transitions, conditional control flow, navigation failures, and checkpoint states. This design provides a unified mechanism for tracing execution, inspecting system states, and implementing recovery strategies.

\section{Related Work}

\subsection{Vision-Language Navigation}

Vision-language navigation (VLN) has evolved from end-to-end methods on discrete navigation graphs to continuous environments that require memory, hierarchical planning, and low-level control. ETPNav maintains an online topological graph of previously visited waypoints and decomposes VLN-CE into high-level topological planning and low-level obstacle avoidance~\cite{an2024etpnav}. MapGPT represents an online topological map in language, enabling a large language model (LLM) to incorporate global spatial structure when planning multistep waypoint candidates~\cite{chen2024mapgpt}. These methods show that compact topological memory can reduce short-sighted decisions based only on local observations and provide useful spatial context for long-horizon planning.

Recent work has also formulated VLN as an explicit reasoning process with LLMs and vision-language models (VLMs). NavGPT performs zero-shot action prediction using observation descriptions, trajectory history, and candidate directions~\cite{zhou2023navgpt}. NavGPT-2 aligns visual features with an LLM and combines them with a navigation policy network, reducing the performance gap between general-purpose models and specialized VLN systems~\cite{zhou2024navgpt2}. DiscussNav coordinates multiple experts for instruction understanding, environmental perception, and task completion estimation, reducing the instability associated with a single self-reasoning process~\cite{long2024discussnav}. Although these methods make navigation decisions more interpretable, their reasoning outputs still need to be grounded in stable and executable motion goals in continuous environments.

Several methods connect language reasoning with continuous control through value maps and constraint-aware planning. VLFM, or Vision-Language Frontier Maps, combines VLM predictions with an occupancy map to construct a language-conditioned frontier value map for zero-shot semantic navigation~\cite{yokoyama2024vlfm}. InstructNav uses a Dynamic Chain-of-Navigation to handle different forms of natural-language navigation instructions and converts language plans into robot trajectories using multi-source value maps~\cite{long2024instructnav}. CA-Nav formulates zero-shot VLN-CE as constraint-aware sub-instruction completion and dynamically updates its plan through constraint state management and cross-modal value maps~\cite{chen2025canav}. However, existing value estimation methods typically operate on images, 2D bounding boxes, or sparse frontiers. These coarse representations provide limited pixel-level semantic grounding, making it difficult to reliably distinguish instruction-relevant targets from background context. Some methods also assume access to depth, panoramic observations, or explicit poses, which does not match the sensing conditions of monocular RGB-only navigation.

Long-horizon tasks and restricted observations pose additional challenges for VLN. LH-VLN introduces a multistage, long-distance navigation benchmark and uses multigranularity dynamic memory to retain information across subtasks~\cite{song2025lhvln}. StreamVLN models navigation as online decision-making over a video stream and balances historical context with inference efficiency through rapidly updated context and slowly updated memory~\cite{wei2025streamvln}. For monocular observations, NaVid formulates VLN as video-to-action prediction~\cite{zhang2024navid}, while MonoDream uses a unified navigation representation and latent panoramic imagination to recover spatial information that is missing from narrow-field-of-view inputs~\cite{wang2026monodream}. Prior work has examined semantic reasoning, mapping, and long-context modeling, but geometric perception and semantic navigation are still commonly handled by separate pipelines with distinct visual features. Long-horizon execution is also often managed through rigid control loops or implicit history representations. To address these limitations, we unify geometric perception and semantic decision-making in a shared dense visual feature space. Combined with a stateful graph orchestration mechanism, this design explicitly manages long-horizon execution for monocular zero-shot VLN-CE.

\subsection{Feed-Forward 3D Reconstruction}

Feed-forward 3D reconstruction estimates depth, point clouds, camera parameters, and other 3D representations from a single image, multiple views, or video in one or a few forward passes. Traditional Structure from Motion (SfM) and Simultaneous Localization and Mapping (SLAM) pipelines rely on feature matching, pose-graph optimization, and iterative refinement for each scene. Recent visual geometry foundation models instead learn from large-scale data with Transformer architectures and generalize more effectively across scenes.

For monocular geometry estimation, Depth Anything V2 improves the robustness and detail of depth prediction through large-scale pseudo-labeling on real images and supervision from synthetic data~\cite{yang2024depthanythingv2}. UniDepth and UniDepthV2 estimate metric 3D structure and camera-related representations from a single image, with a focus on zero-shot generalization across domains~\cite{piccinelli2024unidepth,piccinelli2025unidepthv2}. MoGe formulates open-domain monocular geometry estimation as affine-invariant point-map prediction and jointly recovers depth, point clouds, and field of view~\cite{wang2024moge}. These models reduce the need for active depth sensors in robotic systems. However, frame-wise predictions do not explicitly enforce temporal consistency and therefore cannot guarantee consistent scale and pose across a sequence.

Multiview models further integrate camera estimation, correspondence matching, and dense reconstruction. DUSt3R formulates two-view and multiview reconstruction as point-map regression, predicting dense 3D structure without known camera parameters~\cite{wang2024dust3r}. Building on DUSt3R, MASt3R introduces dense local features and reciprocal matching to improve correspondence estimation under large viewpoint changes~\cite{leroy2024mast3r}. CUT3R maintains a persistent internal state while processing continuous image streams and produces online point maps in a shared coordinate system, making it well suited to the continuous perception requirements of robotic systems~\cite{wang2025cut3r}.

More general models have begun to predict geometry, trajectories, and renderable representations within a unified framework. VGGT, or the Visual Geometry Grounded Transformer, uses a unified Transformer to estimate camera parameters, depth, point maps, and point tracks~\cite{wang2025vggt}. VGGT-SLAM incorporates these outputs into dense RGB SLAM for continuous pose estimation and submap alignment~\cite{maggio2026vggtslam}. AnySplat extends feed-forward reconstruction to 3D Gaussian Splatting for efficient novel-view synthesis~\cite{jiang2025anysplat}, while UniForward combines Gaussian Splatting with open-vocabulary semantic field reconstruction~\cite{tian2025uniforward}. This line of work reflects a shift from isolated depth estimation toward joint modeling of geometry, semantics, and renderable representations. These models provide a natural basis for spatial memories that navigation systems can update online. Our work reuses their visual features for both geometric mapping and language-guided navigation, avoiding redundant visual encoding.

\subsection{Agent Orchestration Frameworks}

Robotic systems have long used finite-state machines, behavior trees, and execution graphs to represent task stages and state transitions. Behavior trees support hierarchical composition and reactive reevaluation, making them well suited to interpretable conditional branches and local recovery procedures~\cite{iovino2022btsurvey}. Execution monitoring methods also address deviation detection, diagnosis, and recovery~\cite{pettersson2005executionmonitoring}, while behavior trees can organize fault-handling actions~\cite{wu2021btrecovery}. Task and Motion Planning (TAMP) jointly considers symbolic task constraints and geometric feasibility~\cite{garrett2021tamp}. However, these methods generally assume predefined skills and explicit task models, and they do not specify how open-vocabulary semantics, online maps, and action histories should be maintained within a shared state.

Embodied language agents combine high-level language planning with environmental feedback and action feasibility. SayCan filters language-model-generated steps according to skill affordances~\cite{ahn2023saycan}. Inner Monologue iteratively revises plans based on success signals and scene feedback~\cite{huang2023innermonologue}. PaLM-E incorporates continuous visual observations into an embodied multimodal language model~\cite{driess2023palme}. Research on active perception further suggests that sensing frequency and observation actions should depend on task requirements~\cite{bajcsy2018activeperception}. These studies address language planning, feedback-based correction, and task-driven perception, but pay less attention to how geometric maps, semantic evidence, progress states, and action histories are jointly maintained over long-horizon navigation.

LLM-based agent systems typically organize functional modules through roles, messages, and task dependencies. MetaGPT encodes standard operating procedures as collaboration protocols among roles~\cite{hong2024metagpt}. AutoGen represents language models, human participants, and tools as customizable conversational components~\cite{wu2023autogen}. GPTSwarm models a multi-agent system as an optimizable node--edge computation graph~\cite{zhuge2024gptswarm}. These frameworks provide general abstractions for modular reasoning and task allocation, but their internal states rarely correspond directly to action outcomes, spatial constraints, or environmental changes during robotic execution.

At the runtime level, ExoFlow supports persistent execution and failure recovery for Directed Acyclic Graph (DAG) workflows~\cite{zhuang2023exoflow}, while LangGraph provides a graph-based abstraction for stateful, multi-participant LLM applications~\cite{langgraph}. Their checkpointing and state reentry mechanisms provide useful design references for long-running workflows. However, the fault-tolerance mechanisms of general-purpose workflow systems do not directly address the needs of embodied navigation. Continuous VLN instead requires an auditable execution graph that supports environment-driven state updates, conditional transitions, checkpoint recovery, and task-level replanning within a unified runtime.

\section{Method}
\label{sec:method}

\subsection{Problem Formulation and Spatial Representation}
\label{sec:formulation}

\subsubsection{Task Definition}

We consider zero-shot Vision-and-Language Navigation in Continuous Environments (VLN-CE). Given a natural-language instruction $I$, the agent receives a monocular RGB observation $\mathbf I_t \in \mathbb R^{H\times W\times 3}$ at decision step $t$ and selects an action $a_t\in\mathcal{A}$, where $\mathcal{A}=\{\alpha_{\mathrm F},\alpha_{\mathrm L},\alpha_{\mathrm R},\alpha_0\}$ corresponds to moving forward, turning left, turning right, and stopping, respectively. Forward and turning actions use fixed translational and angular increments.

Unlike conventional VLN-CE systems that rely on simulator-provided depth or navigation-specific training, our agent operates without depth observations, prior maps, or task-specific navigation training. All geometric information needed for mapping and planning is inferred online from monocular RGB observations. After each motion, we also compare the displacement implied by the issued action with that estimated from visual geometry. This motion-consistency signal is used to detect unreliable traversability estimates and navigation failures.

\subsubsection{Spatial Representation}

We maintain a world coordinate frame $\mathcal W$, a body frame $\mathcal B_t$, a camera frame $\mathcal C_t$, and an agent-centered local map frame $\mathcal L_t$. The agent pose at decision step $t$ is represented by the body-to-world transformation $\mathcal T_{\mathcal B_t}^{\mathcal W}$, which specifies the orientation and position of the body frame $\mathcal B_t$ with respect to the world frame $\mathcal W$. The world frame is initialized from the first reliable geometric observation, with its vertical axis aligned with the estimated ground normal. The horizontal axes of the local map remain aligned with $\mathcal W$, while its origin moves with the agent. Thus, the map orientation remains fixed as the agent changes its heading.

The transformation from the current camera frame to the world frame is denoted by $\mathcal T_{\mathcal C_t}^{\mathcal W}$. Since monocular reconstruction may exhibit small frame-to-frame scale variations even after approximate metric initialization, inter-frame registration is modeled as a bounded similarity transformation in $\mathrm{Sim}(3)$, with the scale factor restricted to a predefined interval. Detailed conversions among camera, body, pixel, and map coordinates are provided in Appendix~\ref{app:coordinate_transform}.

\subsubsection{Global Map and Local Window}
\label{subsec:map_convention}

Navigation relies on two complementary spatial representations. An extensible global bird's-eye-view map, referred to as the {Global Top-Down Map} (GTM) and denoted by $\mathbf G_t$, stores geometric and semantic information accumulated throughout the episode, while a fixed-size local window centered on the agent is used for waypoint generation and short-range planning. Both representations are aligned with $\mathcal W$.

At each decision step, observations are first reconstructed in the current camera frame, transformed into the world frame, and then projected onto the local map. The updated local state is subsequently fused into the GTM. This design preserves long-term spatial memory while keeping local planning computationally tractable.

\subsection{System Framework}
\label{sec:overview}

\begin{figure}[htbp]
\centering
\includegraphics[width=\linewidth]{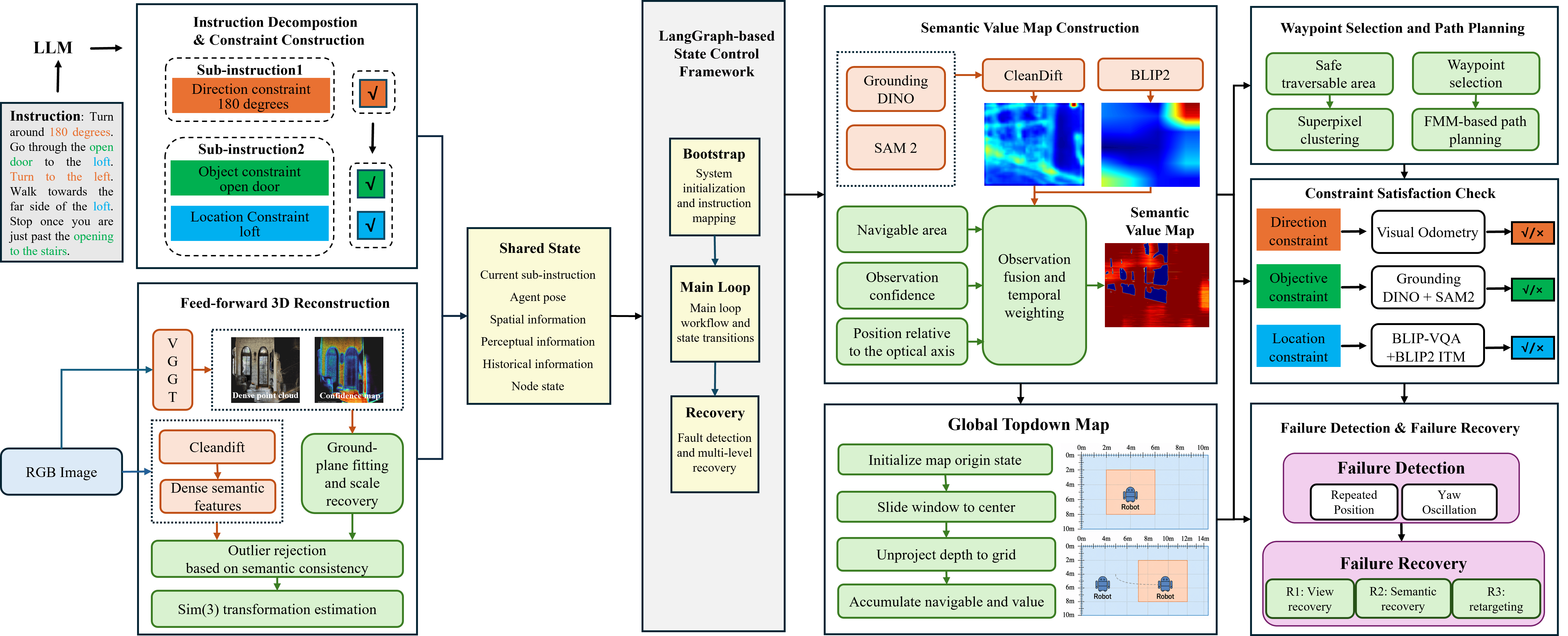}
\caption{Overview of the proposed monocular zero-shot VLN-CE framework.}
\label{fig:framework}
\end{figure}

We propose a monocular zero-shot VLN-CE framework that converts natural-language instructions into executable navigation through four interacting components: instruction constraint tracking, semantic-geometric reconstruction, world-aligned semantic mapping, and hierarchical navigation. A stateful execution graph implemented with LangGraph coordinates these components and maintains persistent navigation state across perception, planning, action execution, and recovery. Fig.~\ref{fig:framework} provides an overview of the framework.

At the beginning of each episode, the instruction is decomposed into an ordered sequence of sub-instructions. Each sub-instruction is associated with explicit completion constraints that describe relevant objects, locations, turns, and relative movement directions. These constraints define both the semantic targets for navigation and the criteria used to assess navigation progress.

For spatial perception, VGGT reconstructs monocular geometry from incoming RGB observations~\cite{wang2025vggt}. Dense semantic features from CleanDIFT~\cite{stracke2025cleandift}, a denoised variant of diffusion-based correspondence features~\cite{tang2023dift}, are used to reject semantically inconsistent geometric correspondences before inter-frame registration. The reconstructed geometry is then transformed into a world-aligned bird's-eye-view map that accumulates traversability, occupancy, exploration, and semantic-value information.

Semantic guidance comes from two complementary visual signals. CleanDIFT provides dense correspondence-based visual relevance, while BLIP-2 estimates image--text relevance for the active sub-instruction. The two signals are fused based on observation confidence and accumulated in the global value map.

Navigation follows a hierarchical design. The high-level planner generates reachable semantic waypoint candidates from the local map and selects a target based on instruction relevance, navigation history, and temporal consistency. The low-level planner then uses the Fast Marching Method (FMM)~\cite{sethian1996fmm} to compute a collision-free short-term goal and converts it into discrete motion commands.

The system operates in a closed loop, with each new observation updating the geometry, semantic evidence, instruction progress, and navigation state. When the execution graph detects sustained lack of progress, repeated visits, or heading oscillation, control is transferred to a recovery subgraph that revises perception or the navigation target before returning to normal execution.

\subsection{Stateful Navigation Graph with LangGraph}
\label{sec:graph_framework}

A central component of our framework is a stateful navigation graph that explicitly models the interactions among perception, reasoning, planning, execution, and recovery. Rather than running these modules as a fixed sequential pipeline, LangGraph maintains a persistent shared state and dynamically routes execution based on navigation progress and failure conditions.

\subsubsection{Stateful Execution Model}
\label{subsec:graph_state}

At step $t$, the shared state contains the active sub-instruction, current pose, global map, navigation history, and graph-level control status. We denote this state compactly by $\mathbf s_t$. Given the current observation $o_t$, graph execution is written as
\begin{equation}
n_t = \rho(\mathbf s_t), \qquad
\mathbf s_t^{+} = F_{n_t}(\mathbf s_t,o_t),
\label{eq:graph_transition}
\end{equation}
where $\rho$ selects the active graph branch and $F_{n_t}$ denotes the state transition performed by that branch. The observation $o_t$ contains the current RGB frame together with the outcome of the preceding action.

The parent graph contains four semantic branches: initialization, normal navigation, recovery, and termination. These branches share the same persistent geometric map and navigation history, allowing recovery to revise the current decision process without discarding spatial information accumulated before the failure.

This graph structure serves two purposes. It separates high-level execution logic from individual perception and planning modules, allowing each module to operate on a consistent navigation state. It also makes non-monotonic navigation behavior explicit: the agent can return from planning to perception, temporarily leave the main navigation loop for recovery, and then resume the interrupted sub-instruction without restarting the episode.

\subsubsection{Initialization Subgraph}
\label{subsec:init}

The initialization subgraph prepares the language, geometric, and orientation states needed for subsequent navigation. It first decomposes the instruction and aligns open-vocabulary entities with the representations used for object detection and visual retrieval. It then estimates the metric reconstruction scale from ground observations and performs an initial panoramic scan.

The panoramic scan collects complementary geometric and semantic observations from multiple headings. For sub-instructions without an explicit turning requirement, the agent uses the accumulated semantic evidence to select a favorable initial orientation. If the instruction explicitly requires a turn, the specified direction takes precedence and remains an unsatisfied directional constraint rather than being overridden by semantic orientation selection.

Initialization is complete only when instruction parsing, semantic alignment, metric scale recovery, and the initial observation scan have all produced valid states. The agent then proceeds with incremental mapping and waypoint planning. Detailed scale-validity criteria and initialization parameters are provided in Appendix~\ref{app:slam_details}.

\subsubsection{Closed-Loop Navigation}
\label{subsec:graph_mainloop}

After initialization, the agent repeatedly performs geometric updates, semantic perception, constraint evaluation, waypoint selection, local planning, and action execution.

Geometric reconstruction and pose estimation are updated from every valid observation because they define the metric state used for mapping and control. More expensive open-vocabulary semantic perception is triggered only when needed. A new semantic observation is requested when the active sub-instruction changes, after recovery, or when the viewpoint has changed sufficiently to provide new task-relevant evidence. Otherwise, the most recent semantic observation is retained in the graph state. The detailed scheduling policy is provided in Appendix~\ref{app:perception_failure}.

After perception, the new geometric and semantic observations are projected into the shared world-aligned map. The instruction module then evaluates the completion constraints associated with the active sub-instruction. Once all constraints are satisfied, the graph advances to the next sub-instruction, suppresses stale target-specific semantic evidence, and requests a fresh semantic observation for the new navigation stage.

If the active sub-instruction remains incomplete, the updated semantic map is used to generate candidate waypoints. A high-level waypoint is selected from the reachable region, and FMM then computes a short-term geometric target. The corresponding discrete action is executed, and its outcome is written back to the graph state, completing the perception--planning--action loop.

\subsubsection{Navigation Failure Detection}
\label{subsec:failure_detection}

Continuous navigation can fail even when a locally valid path exists. Common failure modes include repeated attempts to traverse an obstacle that has been incorrectly classified as traversable, negligible displacement after forward actions, repeated visits to the same neighborhood, oscillation between similar waypoints, and persistent turning without spatial progress.

We detect these conditions using the estimated trajectory, recent actions, waypoint progress, and motion-consistency observations. Rather than treating a single unsuccessful action as a failure, the graph detects persistent loss of progress over a temporal window. This avoids interrupting normal navigation because of transient reconstruction errors or isolated collisions. Detailed detection criteria and thresholds are provided in Appendix~\ref{app:perception_failure}.

Once persistent failure is detected, the current waypoint is suspended and control is transferred to the recovery subgraph.

\subsubsection{Recovery Subgraph}
\label{subsec:recovery_policy}

The recovery subgraph progressively increases the extent to which the current navigation hypothesis is revised. It consists of three recovery levels.

\paragraph{Viewpoint Recovery.}
The agent first performs a panoramic observation at its current location and reevaluates target localization and instruction relevance across the newly observed headings. This stage addresses failures caused by a limited field of view, temporary occlusion, or an unfavorable camera orientation. If reliable task-relevant evidence is recovered, a new waypoint is generated and the graph returns directly to the main navigation loop.

\paragraph{Semantic Recovery.}
If viewpoint recovery remains inconclusive, the graph invalidates target-specific semantic observations associated with the current hypothesis and reconstructs them from the latest visual evidence. Object localization, visual references, and instruction-conditioned relevance are then reevaluated while the persistent geometric map is retained.

\paragraph{Navigation Redirection.}
When the active semantic target does not provide a reliable reachable waypoint, the graph temporarily shifts to geometric exploration. A reachable frontier or unexplored location within the safe connected region is selected as an intermediate target. This target is used only to obtain a new viewpoint and expand the observable space; it does not replace the active instruction constraint.

After successful recovery, temporary planning state is cleared, while the global geometric map, semantic memory, and long-term navigation history are preserved. Semantic perception is refreshed before the graph resumes the interrupted sub-instruction. If repeated recovery attempts remain unsuccessful, the episode is eventually marked as unrecoverable. Additional recovery details are provided in Appendix~\ref{app:recovery_details}.

\subsection{Semantics-Enhanced Monocular 3D Reconstruction}
\label{sec:slam}

Accurate spatial reasoning is challenging in VLN-CE when only monocular RGB observations are available. We therefore construct an online metric representation by combining feed-forward 3D reconstruction with semantics-aware correspondence filtering. VGGT provides dense geometric predictions, while CleanDIFT filters out visually plausible but semantically inconsistent inter-frame matches.

\subsubsection{Feed-Forward Reconstruction and Metric Scale}
\label{subsec:slam_metric}

For each RGB observation $\mathbf I_t$, VGGT predicts dense relative depth, a 3D point map, and geometric confidence~\cite{wang2025vggt}. Since monocular reconstruction does not inherently determine metric scale, we estimate an initial scale using the known camera mounting height $h_{\mathrm{cam}}$.

During initialization, high-confidence points on the observed ground are used to fit a local ground plane. Let $\hat{\mathbf n}_t$ and $\hat b_t$ denote the unit normal and offset of the fitted plane, respectively. Since the plane normal is normalized, $|\hat b_t|$ gives the camera-to-ground-plane distance in the relative reconstruction space. The resulting per-frame scale estimate is
\begin{equation}
\hat s_t
=
\frac{h_{\mathrm{cam}}}{|\hat b_t|}.
\label{eq:height_scale}
\end{equation}

During the initialization scan, observations that satisfy the ground-support, plane-fitting, and scale-validity criteria are retained. We denote the corresponding set of valid frames by $\mathcal F_{\mathrm{valid}}$ and estimate the common initial metric scale robustly as
\begin{equation}
s_0 =
\operatorname{median}
\left(
\left\{
\hat s_t
\mid
t\in\mathcal F_{\mathrm{valid}}
\right\}
\right).
\label{eq:initial_scale}
\end{equation}

The predicted relative depth and 3D point maps are then rescaled by $s_0$ to establish a common approximate metric reference for mapping and navigation. Using multiple initial headings makes the scale estimate less sensitive to local occlusions, ground-segmentation errors, and degenerate plane observations. Detailed ground-plane fitting, frame-validity criteria, and initialization procedures are provided in Appendix~\ref{app:slam_details}.

\subsubsection{Semantic Outlier Rejection with CleanDIFT}
\label{subsec:slam_cleandift}

For frames used in inter-frame registration, CleanDIFT extracts dense semantic features from the input images. The resulting feature maps are interpolated to the input-image resolution and $\ell_2$-normalized along the feature dimension, yielding a unit semantic descriptor $\mathbf f_t(\mathbf u)$ at each pixel $\mathbf u$.

For each candidate 3D correspondence generated by the geometric matching module between consecutive frames, the system measures semantic consistency using the CleanDIFT descriptors at the corresponding image locations. Correspondences with low semantic similarity are rejected, and the remaining correspondences are further filtered using the geometric confidence predicted by VGGT.

This procedure combines geometric reliability from VGGT with semantic consistency from CleanDIFT. Geometric confidence removes points with unreliable reconstruction, while semantic consistency helps reject geometrically plausible correspondences between semantically different regions, particularly in repetitive structures and weak-texture areas. 

\subsubsection{Inter-Frame Similarity Transformation Estimation and Global Fusion}
\label{subsec:slam_sim3}

Our backend optimization follows the general design of~\cite{maggio2026vggtslam}. Although metric scale is initialized during the initialization stage, frame-wise feed-forward reconstruction may still exhibit residual scale variations. We therefore model inter-frame registration as a constrained $\mathrm{Sim}(3)$ transformation. For a geometrically consistent 3D correspondence, points in consecutive frames are expected to satisfy
\begin{equation}
\mathbf x_i^{t}
\simeq
s_{t-1,t}
\mathbf R_{t-1,t}
\mathbf x_i^{t-1}
+
\mathbf t_{t-1,t},
\qquad
\left|
\log s_{t-1,t}
\right|
\le\epsilon_s,
\label{eq:sim3_correspondence}
\end{equation}
where $\mathbf x_i^{t-1}$ and $\mathbf x_i^{t}$ denote corresponding 3D points in the previous and current camera frames, respectively, $s_{t-1,t}>0$ is the inter-frame scale factor, $\mathbf R_{t-1,t}\in\mathrm{SO}(3)$ is the relative rotation, and $\mathbf t_{t-1,t}\in\mathbb R^3$ is the relative translation. The constraint $\left|\log s_{t-1,t}\right|\le\epsilon_s$ bounds frame-to-frame scale variation and prevents a single registration step from introducing an excessive scale change.

The semantically and geometrically filtered correspondences are used to robustly estimate the constrained similarity transformation. RANSAC~\cite{fischler1981ransac} provides an initial estimate while rejecting geometrically inconsistent matches, after which the transformation parameters are refined with a confidence-weighted robust objective. Details of correspondence construction, semantic filtering, RANSAC estimation, correspondence weighting, and robust $\mathrm{Sim}(3)$ refinement are provided in Appendix~\ref{app:slam_details}.

The estimated inter-frame transformation is then used to update the current camera pose and register the reconstructed geometry in the world-aligned map. If a reliable registration cannot be obtained, the previous pose estimate is retained and global map fusion is temporarily suspended. Detailed failure-handling and map-fusion criteria are provided in Appendix~\ref{app:slam_details}.

\subsection{Instruction Decomposition and Constraint Tracking}
\label{sec:csm}

\subsubsection{Semantics-Guided Sub-instruction Decomposition}
\label{sec:instruction_decomposition}

VLN instructions often describe a sequence of spatially distinct operations. Reasoning over the full instruction at every step can introduce landmarks or actions that are irrelevant to the current navigation stage, potentially interfering with visual grounding~\cite{ideal_vln}. We therefore represent each instruction as an ordered sequence of sub-instructions, each associated with an explicit set of completion constraints.

Candidate boundaries are identified using navigation verbs, sequential connectives, changes in spatial relations, and syntactic structure. A lightweight LLM then normalizes the resulting segments and extracts task-relevant entities and relations. This hybrid decomposition limits reliance on unconstrained language generation while retaining the flexibility needed to handle open-vocabulary instructions.

For each sub-instruction $I_k$, we construct a constraint set $\mathcal C_k$ containing object, location, and direction constraints as needed. Explicit turning commands and relative movement directions are represented separately: turning commands specify a change in orientation, whereas relative movement directions specify displacement with respect to the heading at the start of the sub-instruction. Instructions that require backward motion are normalized into a turn followed by forward movement, allowing the low-level action space to remain unchanged.

\subsubsection{Sub-instruction Constraint Satisfaction}
\label{subsec:csm_sat}

At decision step $t$, the graph evaluates all constraints associated with the active sub-instruction $I_{k_t}$. The completion state is defined as
\begin{equation}
\chi_t^{\mathrm{done}}
=
\prod_{c\in\mathcal C_{k_t}}
\operatorname{Sat}_t(c),
\label{eq:subinstruction_done}
\end{equation}
where $\operatorname{Sat}_t(c)\in \{0,1\}$ indicates whether constraint $c$ is supported by the visual observations, estimated trajectory, and action history available up to step $t$. The graph proceeds to the next sub-instruction only after all active constraints are satisfied.

\paragraph{Object Constraints.}
For open-vocabulary landmarks, we use Grounding DINO to obtain language-conditioned detections~\cite{liu2023groundingdino}, followed by SAM~2~\cite{ravi2024sam2} to refine the corresponding instance regions. Valid pixels in each mask are associated with reconstructed depth values and projected into the world frame. The target position is robustly estimated from the resulting 3D points and updated as new semantic evidence becomes available. An object constraint is considered satisfied when the target has been observed with sufficient confidence and the agent has reached the required spatial relationship to it.

\paragraph{Location Constraints.}
\label{subsec:csm_location}

Location constraints represent transitions between semantic regions rather than individual visual detections. We use visual question answering to determine whether the current observation is consistent with a target region, with BLIP-2 image--text relevance used as a fallback when needed~\cite{li2023blip2}. Rather than classifying each frame independently, the graph maintains a short-term region state that records whether the agent is entering, remaining inside, leaving, has previously visited, or is traversing a region. A traversal constraint is therefore satisfied only by a consistent sequence of entry, presence, and exit events, rather than by a single positive classification. The full state-transition rules are given in Appendix~\ref{app:constraint_details}.

\paragraph{Direction Constraints.}
\label{subsec:csm_direction}

For each sub-instruction, the agent's heading at activation is recorded as the reference orientation. Turning constraints are evaluated using the accumulated heading change relative to this reference. Relative movement-direction constraints, by contrast, are evaluated from the direction of the estimated displacement over a short temporal window. Separating these two cases prevents an in-place rotation from satisfying instructions such as ``move to the left'' and prevents lateral motion from being interpreted as an explicit ``turn left'' command. The angular ranges and temporal parameters used for different direction constraints are reported in Appendix~\ref{app:constraint_details}.

Once a sub-instruction is completed, the graph advances to the next sub-instruction, downweights semantic values associated with the previous target, and refreshes target-specific visual evidence. The geometric exploration history is retained across this transition.

\subsection{Constraint-Guided Semantic Value Estimation}
\label{sec:value_estimation}

The active instruction constraints are transformed into a spatial signal for waypoint selection. We obtain this signal by integrating dense visual correspondence with image--text relevance, which provide complementary semantic cues.

Let $\mathbf A_t$ denote the CleanDIFT-based relevance map between the current observation and the visual references associated with the active constraint, and let $\mathbf B_t$ denote the BLIP-2 image--text relevance map. The available semantic observations are fused as
\begin{equation}
\mathbf E_t =
\frac{
\beta_a q_t^a \mathbf A_t
+
\beta_b q_t^b \mathbf B_t
}{
\beta_a q_t^a
+
\beta_b q_t^b
+
\epsilon
},
\label{eq:semantic_evidence_fusion}
\end{equation}
where $q_t^a$ and $q_t^b$ represent the reliability of the two semantic sources, and $\beta_a,\beta_b$ determine their relative contributions. If one source is unavailable, the remaining source is used to construct the evidence map.

The semantic evidence is then weighted by observation confidence. This confidence accounts for geometric validity, viewing direction, observation range, surface geometry, traversability, and pose reliability. Specifically, observations close to the optical axis and locations supported by reliable reconstruction and registration are assigned higher weights, whereas obstacle regions and geometrically uncertain pixels are down-weighted.

To improve temporal consistency, historical semantic observations are reprojected into the current image using the estimated inter-frame transformation. Let $\widetilde{\mathbf V}_t^{\mathrm{img}}$ and $\widetilde{\mathbf Q}_t^{\mathrm{val}}$ denote the current-frame value and confidence maps, respectively, and let $\widehat{\mathbf V}_{t-1\rightarrow t}$ and $\widehat{\mathbf Q}_{t-1\rightarrow t}$ denote the corresponding historical estimates after reprojection into the current view.

The navigation mask $\mathbf M_t^{\mathrm{nav}}$ first excludes non-traversable
pixels and pixels occluded by foreground obstacles. Reprojections within the
remaining valid regions are further filtered by image-boundary, positive-depth,
and occlusion-consistency constraints, with the resulting validity encoded by
$\mathbf M_t^{\mathrm{rep}}$. The current and historical observations are then
combined through confidence-weighted averaging:

\begin{equation}
\mathbf V_t^{\mathrm{img}} =
\mathbf M_t^{\mathrm{nav}}
\odot
\frac{
\widetilde{\mathbf Q}_t^{\mathrm{val}}
\odot
\widetilde{\mathbf V}_t^{\mathrm{img}}
+
\eta \,
\mathbf M_t^{\mathrm{rep}}
\odot
\widehat{\mathbf Q}_{t-1\rightarrow t}
\odot
\widehat{\mathbf V}_{t-1\rightarrow t}
}{
\widetilde{\mathbf Q}_t^{\mathrm{val}}
+
\eta \,
\mathbf M_t^{\mathrm{rep}}
\odot
\widehat{\mathbf Q}_{t-1\rightarrow t}
+
\epsilon
},
\label{eq:image_value_temporal}
\end{equation}
and the corresponding confidence map is updated as
\begin{equation}
\mathbf Q_t^{\mathrm{val}} =
\mathbf M_t^{\mathrm{nav}}
\odot
\operatorname{clip}
\left(
\widetilde{\mathbf Q}_t^{\mathrm{val}}
+
\eta \,
\mathbf M_t^{\mathrm{rep}}
\odot
\widehat{\mathbf Q}_{t-1\rightarrow t},
\,
0,
\,
1
\right).
\label{eq:image_conf_temporal}
\end{equation}
Here, $\eta\in[0,1]$ controls the contribution decay of historical observations. Additional implementation details regarding camera-axis confidence, reprojection validity, and local-block interpolation are provided in Appendix~\ref{app:value_details}.

\subsection{Global Bird's-Eye-View Semantic-Geometric Map}
\label{sec:gtm}

The global map $\mathbf G_t$ provides a shared representation for geometric reconstruction, semantic grounding, and navigation planning. It contains traversability, occupancy, exploration, observation-count, semantic-value, and dense semantic-feature layers. The exploration state is derived from accumulated observations, while semantic values and features are associated with the active instruction.

For each observation, reconstructed 3D points are transformed by $\mathcal T^{\mathcal W}_{\mathcal C_t}$ and rasterized into the agent-centered local map. Ground-supported points contribute to the traversability layer, whereas elevated structures contribute to occupancy. When conflicting evidence falls within the same cell, occupancy takes precedence to maintain conservative navigation.

The motion-consistency signal provides an additional source of geometric supervision. If the agent issues a forward command but the estimated pose shows substantially less displacement than expected, cells immediately along the attempted direction are marked as potentially occupied. This feedback allows the map to correct visually inferred traversable regions that repeatedly prove impassable during execution.

The semantic-value and semantic-feature layers are projected using the same coordinate transformation. Repeated observations of the same grid cell are fused according to observation confidence, assigning greater weight to geometrically reliable measurements obtained from favorable viewpoints. Semantic features are re-normalized after fusion. When the active sub-instruction changes, semantic values associated with the previous target are attenuated rather than discarded, preserving useful exploration context while giving priority to the new target.

A fixed-size local window is extracted from the global map for waypoint selection and FMM planning. Further details of map fusion are provided in Appendix~\ref{app:map_planning_details}.

\subsection{Waypoint Selection and Local Planning}
\label{sec:planning}

\subsubsection{Reachable Region and Candidate Generation}
\label{subsec:waypoint}

Waypoint generation is restricted to locations that are both traversable and safely reachable from the current agent position. Let $\Omega_t^{\mathrm{occ}}$ denote the occupied cells in the local map. We define
\begin{equation}
\mathcal R_t =
\operatorname{CC}\left(
\left\{
\mathbf x \;\middle|\;
\mathbf N_t^{\mathrm{nav}}(\mathbf x)=1,\;
\mathbf N_t^{\mathrm{occ}}(\mathbf x)=0,\;
\operatorname{dist}(\mathbf x,\Omega_t^{\mathrm{occ}})
> d_{\mathrm{safe}}
\right\},
\mathbf p_t
\right),
\label{eq:feasible_region}
\end{equation}
where $\operatorname{CC}(\cdot,\mathbf p_t)$ returns the connected component containing the current agent position, and $d_{\mathrm{safe}}$ specifies the obstacle-clearance margin.

Within $\mathcal R_t$, the semantic-value and semantic-feature layers are grouped into spatially coherent regions using a semantics-augmented SLIC clustering~\cite{achanta2012slic}. Each valid region yields a representative waypoint near its geometric center. Regions that are too small, unsafe, unreachable, or strongly associated with a previously detected navigation loop are discarded.

This region-level representation is more robust than selecting individual high-value cells because semantic value maps may contain local peaks caused by viewpoint changes or reconstruction noise. Spatial aggregation provides more stable long-range targets while preserving fine-grained instruction relevance.

\subsubsection{Semantic Waypoint Selection}
\label{subsec:waypoint_stability}

Let $\mathbf V_t^{\mathrm{map}}$ denote the map-level semantic value layer. Let $\mathcal A_t^{\mathrm{wp}}$ denote the remaining candidate waypoints. The semantic score of each candidate $\mathbf w$ is computed by averaging $\mathbf V_t^{\mathrm{map}}$ over its associated region $\mathcal U(\mathbf w)$. The highest-scoring candidate is

\begin{equation}
J_t(\mathbf w)
=
\frac{1}{|\mathcal U(\mathbf w)|}
\sum_{\mathbf x\in\mathcal U(\mathbf w)}
\mathbf V_t^{\mathrm{map}}(\mathbf x),
\qquad
\widehat{\mathbf w}_t
=
\arg\max_{\mathbf w\in\mathcal A_t^{\mathrm{wp}}}
J_t(\mathbf w).
\label{eq:new_waypoint_best}
\end{equation}

Replacing the current waypoint whenever a new candidate receives a slightly higher score can lead to oscillatory behavior. We therefore apply waypoint hysteresis: the previous target is retained as long as it remains safely reachable, has not yet been reached, and its updated semantic score remains close to that of the best newly generated candidate. A new target is selected only when the previous waypoint becomes invalid, is reached, loses substantial semantic relevance, or is associated with a navigation failure.

The graph also stores waypoint history in the world frame. When the agent revisits a previously explored neighborhood, candidates that would repeat an earlier unsuccessful decision are suppressed. If loop suppression removes all candidates, the planner falls back to the highest-valued safe candidate available before history-based filtering rather than terminating navigation.

During the final navigation stage, a confidently detected target can directly refine the stopping waypoint. Its instance mask is combined with valid reconstructed depth to estimate the 3D target position, whose ground-plane projection is matched to the nearest reachable map cell. This target-specific refinement is applied only when sufficient geometric and semantic evidence is available.

\subsubsection{FMM Planning and Action Execution}
\label{subsec:fmm}

The selected high-level waypoint is projected onto the current local map and supplied to FMM, which computes a geodesic distance field over $\mathcal R_t$. A short-term goal is then selected within the reachable local region by following the descending direction of this field for a fixed planning horizon.

The low-level controller compares the direction to the short-term goal with the current agent heading. If the angular difference is within the controller tolerance, the agent executes $\alpha_{\mathrm F}$; otherwise, it executes $\alpha_{\mathrm L}$ or $\alpha_{\mathrm R}$ according to the sign of the angular error. Reaching an intermediate waypoint triggers waypoint regeneration rather than a transition to the next sub-instruction.

Stopping is handled separately from waypoint arrival. The stop action is issued only when the final sub-instruction is complete and the agent has reached the required neighborhood of the final target. The navigation episode terminates when the stop action is executed or when the predefined action budget is exhausted.

\section{Experiments}
\label{sec:exp}

\subsection{Experimental Setup}
\label{subsec:simulation}

We conduct all simulation experiments in Habitat-Sim~\cite{savva2019habitat} using scenes from the Matterport3D (MP3D) dataset~\cite{chang2017matterport}. All experiments run on a system equipped with a single NVIDIA RTX 5090 GPU with 32 GB of memory, an Intel Xeon Platinum 8470Q CPU, and 90 GB of RAM, running Ubuntu 22.04. We follow the standard R2R-CE evaluation protocol~\cite{anderson2018vln,krantz2020vlnce} and evaluate on the val-unseen split to assess the agent's ability to generalize to unseen indoor environments.

We use a fixed evaluation subset constructed from the R2R-CE val-unseen split. Specifically, we sample 50 navigation episodes from each of the 11 unseen scenes, resulting in 550 episodes in total. The same subset is used throughout the experiments, including the main comparisons, ablation studies, and analyses of navigation efficiency, depth estimation and 3D reconstruction, and pose estimation.

For the main comparisons, we reproduce each method on this subset using its publicly released configuration. All methods are evaluated in the same Habitat-Sim environment with identical task definitions, action spaces, and evaluation metrics. Using the same episodes and evaluation setup across methods ensures a consistent comparison while keeping the computational cost of repeated evaluations manageable.

We do not use data from the R2R-CE training split to train the navigation policy or construct task-specific data for fine-tuning. The language model and visual foundation models in our system rely on publicly available pretrained weights and are used only for instruction understanding and scene-level semantic reasoning. The agent receives only first-person monocular RGB images as visual input and does not use auxiliary information from Habitat, such as ground-truth depth, semantic segmentation, ground-truth poses, or global maps. Instead, the system estimates the depth, camera poses, and map representations required for navigation online.

The agent operates in a low-level discrete action space consisting of \textsc{Forward}, \textsc{TurnLeft}, \textsc{TurnRight}, and \textsc{Stop}. A \textsc{Forward} action moves the agent by $0.25\ \mathrm{m}$, while each \textsc{TurnLeft} or \textsc{TurnRight} action rotates it by $30^\circ$. Before navigation begins, the agent first obtains a panoramic observation, either using a panoramic camera or by rotating a monocular camera through a full $360^\circ$. The resulting panoramic observations are then used for metric-scale estimation and initial map construction. This initialization stage is treated as a pre-navigation sensing procedure and is excluded from the navigation action budget and evaluation trajectory. The navigation episode starts from the agent state reached after this initialization stage. From this post-initialization state, all methods are evaluated on the same 550-episode subset using an identical budget of 150 low-level navigation actions per episode.

We evaluate navigation performance using four standard R2R-CE metrics: Success Rate (SR), Oracle Success Rate (OSR), Success weighted by Path Length (SPL), and Navigation Error (NE). We use a success threshold of $3.0\ \mathrm{m}$. SR measures the proportion of episodes in which the agent stops within $3.0\ \mathrm{m}$ of the goal. OSR measures the proportion of episodes whose trajectories reach at least one position within $3.0\ \mathrm{m}$ of the goal, regardless of where the agent ultimately stops. SPL combines success with the ratio of the shortest-path length to the executed trajectory length, capturing navigation efficiency. NE is the average distance from the agent's final position to the goal. Higher SR, OSR, and SPL indicate better performance, while lower NE indicates more accurate final stopping. All metrics are computed from the navigation trajectory after the initial scale calibration stage; the trajectory generated during calibration is excluded.

\subsection{Experimental Results}

Table~\ref{tab:main_results} compares LG-VLN with four representative zero-shot VLN methods, including NavGPT, Open-Nav, InstructNav, and CA-Nav. NavGPT is evaluated on a preconstructed discrete navigation graph, whereas Open-Nav, InstructNav, and CA-Nav are evaluated in continuous environments. Since these methods differ in their sensor configurations, visual inputs, navigation settings, and action spaces, we also report their key system settings to clarify the conditions under which the results are obtained.

NavGPT~\cite{zhou2023navgpt} converts visual observations at candidate nodes into natural-language descriptions, representing both scene information and candidate-node attributes as text. It performs zero-shot navigation on a preconstructed discrete navigation graph (Nav-graph). Open-Nav~\cite{qiao2024opennav} uses panoramic RGB-D observations and a waypoint prediction module, with an open-source language model performing step-by-step navigation reasoning. InstructNav~\cite{long2024instructnav} performs zero-shot path planning using multi-source value maps; in its egocentric setting, it relies on RGB-D observations and ground-truth poses. CA-Nav~\cite{chen2025canav} takes egocentric RGB-D images and odometry as input and generates navigation waypoints through constraint-aware sub-instruction management and value-map updates. In contrast, LG-VLN uses only egocentric monocular RGB images and employs VGGT to estimate depth, camera poses, and the spatial map online during navigation.

The Efficient LLM Usage column indicates whether a method avoids querying the language model at every low-level navigation step. The Task-Specific Training Module column indicates whether a method relies on waypoint predictors, navigation policies, or other task-specific networks trained using navigation data, environment topology, or related task data. General-purpose vision and language foundation models pretrained on generic data and kept frozen during navigation are not counted as task-specific training modules.

\begin{table}[htbp]
\centering
\caption{Main zero-shot VLN-CE results on a fixed 550-episode subset of the R2R-CE val-unseen split; NavGPT performs navigation over a preconstructed discrete navigation graph, whereas the other methods operate in the continuous Habitat-Sim environment.}
\label{tab:main_results}

\small
\setlength{\tabcolsep}{1.2pt}
\renewcommand{\arraystretch}{1.15}

\renewcommand{\tabularxcolumn}[1]{m{#1}}

\begin{tabularx}{\columnwidth}{
@{}
>{\hsize=0.95\hsize\raggedright\arraybackslash}X
>{\hsize=1.05\hsize\centering\arraybackslash}X
>{\centering\arraybackslash}X
>{\centering\arraybackslash}X
>{\centering\arraybackslash}X
>{\centering\arraybackslash}X
c
c
c
c
@{}
}
\toprule

Method &
\makecell[c]{Sensor\\Configuration} &
\makecell[c]{Efficient\\LLM Usage} &
\makecell[c]{Egocentric\\Obs.} &
\makecell[c]{Task-Specific\\Training Module} &
\makecell[c]{Online Geometry\\Estimation} &
SR$\uparrow$ &
SPL$\uparrow$ &
OSR$\uparrow$ &
NE$\downarrow$ \\

\midrule

NavGPT\textsuperscript{\dag}~\cite{zhou2023navgpt} &
\makecell[c]{Panoramic\\RGB} &
$\times$ &
$\times$ &
$\times$ &
$\times$ &
31.8 &
26.8 &
40.9 &
9.75 \\

Open-Nav~\cite{qiao2024opennav} &
\makecell[c]{Panoramic\\RGB-D} &
$\times$ &
$\times$ &
$\checkmark$ &
$\times$ &
26.4 &
23.2 &
31.6 &
7.99 \\

InstructNav~\cite{long2024instructnav} &
\makecell[c]{RGB-D\\+ GT Pose} &
$\times$ &
$\checkmark$ &
$\times$ &
$\times$ &
15.1 &
6.7 &
25.1 &
7.84 \\

CA-Nav~\cite{chen2025canav} &
\makecell[c]{RGB-D\\+ Odometry} &
$\checkmark$ &
$\checkmark$ &
$\times$ &
$\times$ &
22.4 &
9.3 &
43.7 &
8.76 \\

\midrule

LG-VLN &
\makecell[c]{Monocular\\RGB\textsuperscript{\ddag}} &
$\checkmark$ &
$\checkmark$ &
$\times$ &
$\checkmark$ &
21.3 &
12.1 &
29.5 &
7.85 \\

\bottomrule
\end{tabularx}

\vspace{1mm}
\noindent
\parbox{\linewidth}{%
\footnotesize
$\dagger$ NavGPT is evaluated on a preconstructed discrete navigation graph
rather than in the continuous environment. Its reported metrics are therefore
provided for reference only and are not directly comparable with those of the
continuous-environment methods.

$\ddagger$ LG-VLN additionally assumes a known camera mounting height
$h_{\mathrm{cam}}$ for metric-scale initialization.
}

\end{table}

NavGPT and Open-Nav achieve relatively strong SR, reaching 31.8\% and 26.4\%, respectively, along with competitive SPL. NavGPT, however, does not navigate from online sensor observations in a continuous environment. Instead, it plans over a preconstructed discrete navigation graph (Nav-graph), whereas all other methods operate directly in the continuous Habitat-Sim environment. Open-Nav combines panoramic RGB-D observations with a waypoint predictor trained on VLN datasets, benefiting from both panoramic visual information and task-specific navigation priors.

InstructNav, CA-Nav, and LG-VLN all use egocentric observations, which more closely reflect practical robotic navigation settings. CA-Nav and LG-VLN also avoid querying the LLM at every low-level navigation step, reducing the latency caused by repeated language-model inference. Under this setting, LG-VLN achieves an SR of 21.3\% and an SPL of 12.1\% using only monocular RGB input. Its SR is close to that of CA-Nav (22.4\%), which uses RGB-D observations and odometry, and is notably higher than that of InstructNav (15.1\%), which uses ground-truth poses.

Among the methods that operate with egocentric observations in the continuous environment, CA-Nav achieves an OSR of 43.7\%, compared with 29.5\% for LG-VLN, a gap of 14.2 percentage points. LG-VLN, however, achieves a higher SPL of 12.1\%, compared with 9.3\% for CA-Nav. This difference between OSR and SPL is closely related to the methods' exploration and stopping behavior.

OSR only requires the trajectory to reach a position within $3.0\ \mathrm{m}$ of the goal at any point during navigation. It is therefore sensitive to the extent of exploration and whether the trajectory enters the goal vicinity. CA-Nav uses RGB-D observations and real-time odometry, providing stable metric information for occupancy-grid distance estimation and supporting more aggressive frontier selection based on reachability. LG-VLN, in contrast, estimates depth and camera poses online from monocular RGB observations. Its initial metric scale $s_0$ remains fixed after calibration, while bounded inter-frame $\mathrm{Sim}(3)$ registration compensates for residual frame-to-frame scale fluctuations. As a result, LG-VLN adopts a more conservative exploration radius and frontier-selection policy. This reduces the likelihood that its trajectory enters the goal vicinity during exploration and may contribute to the 14.2 percentage-point OSR gap between LG-VLN and CA-Nav.

Once LG-VLN determines that the goal is nearby and enters its recovery or termination procedure, locally consistent relative depth estimates support waypoint selection without substantially increasing the traveled distance. CA-Nav explores more aggressively, increasing the chance of entering the goal vicinity and thereby improving OSR, but this behavior can also produce longer trajectories. The resulting increase in path length lowers SPL and may explain why CA-Nav achieves a lower SPL despite its higher OSR.

The methods also differ in how they obtain geometric information. InstructNav uses RGB-D observations and ground-truth poses, whereas CA-Nav relies on RGB-D observations and real-time odometry. LG-VLN instead estimates depth, camera poses, and the spatial map online. It therefore does not require a depth sensor, odometry, a prior map, or task-specific modules trained on VLN data. These results suggest that online geometric perception and mapping can provide sufficiently reliable spatial information for continuous navigation while reducing dependence on additional sensors. More broadly, they indicate that competitive VLN performance can be achieved using only monocular RGB observations.

\subsection{Ablation Study}
\label{subsec:ablation}

We conduct ablation experiments on the R2R-CE val-unseen evaluation subset to assess the contribution of each module to overall performance. All ablations use the same parameter settings, with only the module under study removed or replaced.

\subsubsection{Analysis of Shared Visual Feature Enhancement}

The semantic value map combines two complementary visual signals: visual reference similarity from CleanDIFT and image--text relevance from BLIP-2. CleanDIFT captures fine-grained correspondences between regions in the current observation and the target visual reference, while BLIP-2 measures the semantic relevance of the current observation to the active sub-instruction. The full model combines these signals to estimate the semantic values of candidate regions.

We vary the fusion weights $\beta_a$ and $\beta_b$ to assess the contribution of each signal. Setting $\beta_a=0$ removes the CleanDIFT term from the semantic value map, leaving only the BLIP-2-based image--text relevance. Conversely, setting $\beta_b=0$ retains only the CleanDIFT-based visual similarity. This ablation isolates the roles of the two signals in constructing the semantic value map; shared visual features used by other components of the framework remain unchanged.

\begin{table*}[htbp]
\centering
\caption{Ablation on the two semantic sources of the value map (550-episode subset of R2R-CE val-unseen).}
\label{tab:ablation}
\small
\vspace{-0.2em}
\begin{tabular*}{\textwidth}{@{\extracolsep{\fill}}cccccc@{}}
\toprule
\textbf{Mode} & \textbf{$\beta_a$ / $\beta_b$} & SR$\uparrow$ & SPL$\uparrow$ & OSR$\uparrow$ & NE$\downarrow$ \\
\midrule
\textbf{Full \method{}} & 1.0 / 1.0 & \textbf{21.3} & \textbf{12.1} & \textbf{29.5} & \textbf{7.85} \\
w/o CleanDIFT value & 0.0 / 1.0 & 3.0 & 2.1 & 7.8 & 8.70 \\
w/o BLIP-2 value& 1.0 / 0.0 & 12.0 & 9.3 & 16.2 & 8.92 \\
\bottomrule
\end{tabular*}
\end{table*}

As shown in Table~\ref{tab:ablation}, the full LG-VLN model achieves an SR of 21.3\%, an SPL of 12.1\%, and an OSR of 29.5\% on the evaluation subset. Removing either semantic signal consistently reduces navigation performance, supporting their complementary roles in semantic value estimation. Without BLIP-2, SR decreases to 12.0\% and OSR to 16.2\%, indicating that image--text relevance provides useful instruction-conditioned semantic information beyond CleanDIFT-based visual correspondence.

The performance degradation is larger when CleanDIFT is removed, with SR dropping to 3.0\%, SPL to 2.1\%, and OSR to 7.8\%. This suggests that fine-grained visual correspondence plays an important role in distinguishing spatially relevant regions in the proposed value estimator, while BLIP-2 alone is insufficient to provide reliable navigation guidance under the current value-map construction and waypoint-selection pipeline. This result should therefore be interpreted within the proposed fusion framework rather than as a general comparison between the standalone capabilities of CleanDIFT and BLIP-2.

NE varies less than SR and OSR across the three configurations, ranging from $7.85$ to $8.92$\,m. This is consistent with the different definitions of these metrics. NE is a continuous measure of the final distance to the goal, whereas SR depends on whether the final position lies within the success threshold and OSR on whether the trajectory enters the success region at any point. Overall, the results support the complementary use of CleanDIFT for fine-grained visual correspondence and BLIP-2 for instruction-conditioned semantic relevance. This ablation evaluates only their contributions to semantic value-map construction; the effect of shared CleanDIFT features on the geometric front-end is analyzed separately.

\subsubsection{Analysis of Online Geometric Estimation}

To evaluate how online geometric estimation affects VLN performance, we conduct ablation experiments on the same evaluation subset under three geometric settings. The Oracle setting uses ground-truth depth and poses provided by Habitat. The Hybrid setting estimates depth online with VGGT while using ground-truth poses from Habitat. In our setting, both depth and poses are estimated online with VGGT and used for online 3D reconstruction.

All three settings use the same navigation parameters, semantic modules, and planning modules, with only the geometric inputs changed. The resulting performance differences therefore primarily reflect the effects of depth and pose estimation quality on navigation.

\begin{table}[h]
\centering
\caption{Effect of Online Geometric Estimation on Navigation Performance}
\label{tab:geometry_ablation}
\small
\setlength{\tabcolsep}{4pt}
\renewcommand{\arraystretch}{1.15}

\begin{tabularx}{\linewidth}{
>{\centering\arraybackslash}X
>{\centering\arraybackslash}X
>{\centering\arraybackslash}X
>{\centering\arraybackslash}X
>{\centering\arraybackslash}X
>{\centering\arraybackslash}X
}
\toprule
\textbf{Mode}
&
\textbf{Depth Source}
&
\textbf{Pose Source}
&
\textbf{SR$\uparrow$}
&
\textbf{SPL$\uparrow$}
&
\textbf{NE$\downarrow$}
\\
\midrule
Oracle
&
GT Depth
&
GT Pose
&
\textbf{24.9}
&
\textbf{17.7}
&
\textbf{6.80}
\\

Hybrid
&
Pred.
&
GT Pose
&
24.2
&
15.2
&
7.04
\\

Ours
&
Pred.
&
Pred.
&
21.3
&
12.1
&
7.85
\\
\bottomrule
\end{tabularx}
\end{table}

Table~\ref{tab:geometry_ablation} reports the results of the three configurations evaluated on the same set of episodes. Navigation performance gradually declines as ground-truth geometric information is replaced by online estimates. From Oracle to Hybrid and Ours, SR decreases from 24.9\% to 24.2\% and 21.3\%, respectively, while SPL decreases from 17.7\% to 15.2\% and 12.1\%. NE increases from $6.80\,\text{m}$ in the Oracle setting to $7.85\,\text{m}$ in our setting. These results indicate that errors in online depth and pose estimation can affect both navigation success and path efficiency.

The gap between Hybrid and Oracle mainly reflects the effect of VGGT depth estimation errors. Replacing ground-truth depth with VGGT estimates reduces SR and SPL by 0.7 and 2.5 percentage points, respectively, while increasing NE by $0.24\,\text{m}$. The reduction in SR is relatively small compared with the performance gaps among the complete navigation systems reported in Table~\ref{tab:main_results}. This suggests that, given continuous-view observations and initial scale calibration, VGGT depth estimates preserve sufficiently accurate relative depth and local geometric structure to support state-machine constraints, value-map accumulation, and waypoint selection.

The gap between Ours and Hybrid mainly reflects the effect of online pose estimation errors. Replacing ground-truth poses with online estimates further reduces SR and SPL by 2.9 and 3.1 percentage points, respectively, while increasing NE by $0.81\,\text{m}$. This increase in NE is substantially larger than that caused by replacing ground-truth depth ($0.81\,\text{m}$ vs.\ $0.24\,\text{m}$). The comparison therefore suggests that pose estimation errors have a greater effect on the final navigation error than depth estimation errors under the current setup.

\subsection{Navigation Efficiency Analysis}
\label{subsec:efficiency}

LG-VLN includes several computationally intensive components, particularly geometric reconstruction and dense semantic feature extraction. We therefore measure the runtime of the main perception and planning modules to quantify the computational cost of the system. All measurements are collected on the same hardware using identical input settings.

\begin{table}[h]
\centering
\caption{Runtime per call for the main perception and planning modules.}
\label{tab:runtime}
\small
\setlength{\tabcolsep}{4pt}
\renewcommand{\arraystretch}{1.12}

\begin{tabularx}{\linewidth}{
>{\centering\arraybackslash}p{0.25\linewidth}
>{\centering\arraybackslash}p{0.25\linewidth}
>{\raggedright\arraybackslash}X
}
\toprule
\textbf{Module}
&
\textbf{Time/Call $\downarrow$}
&
\textbf{Description}
\\
\midrule

VGGT
&
$178\,\text{ms}$
&
Depth, pose, and point cloud estimation, executed for every frame
\\

CleanDIFT
&
$179\,\text{ms}$
&
Dense semantic feature extraction for inter-frame registration
\\

Grounding DINO + SAM2
&
$602\,\text{ms}$
&
Open-vocabulary detection and instance mask generation, executed when perception is triggered
\\

BLIP-2 ITM
&
$52$--$72\,\text{ms}$
&
Image--text matching, executed when perception is triggered
\\

GTM + FMM
&
$86\,\text{ms}$
&
Map fusion and local planning
\\
\bottomrule
\end{tabularx}
\end{table}

As shown in Table~\ref{tab:runtime}, Grounding DINO + SAM2 is the most computationally expensive semantic perception component, requiring approximately $602\,\text{ms}$ per call. VGGT and CleanDIFT require $178\,\text{ms}$ and $179\,\text{ms}$ per call, respectively, resulting in similar computational costs. BLIP-2 ITM requires $52$--$72\,\text{ms}$ for image--text matching for each short-text candidate, while GTM + FMM takes approximately $86\,\text{ms}$ for map fusion and local planning.

LG-VLN does not execute every semantic model at each navigation step. In the current implementation, VGGT and CleanDIFT are run on every frame to support geometric reconstruction and inter-frame feature association. Grounding DINO + SAM2 and BLIP-2 ITM are invoked only when a semantic perception update is needed. Detection, segmentation, and image--text matching are repeated when the target category changes, the active sub-instruction switches, or the current observation satisfies the perception-update criterion. For consecutive observations within the same sub-instruction, the system reuses existing perception results when no update is required, avoiding redundant computation.

\subsection{Analysis of the Navigation Process}

\begin{figure}[htbp]
\centering
\includegraphics[width=\linewidth]{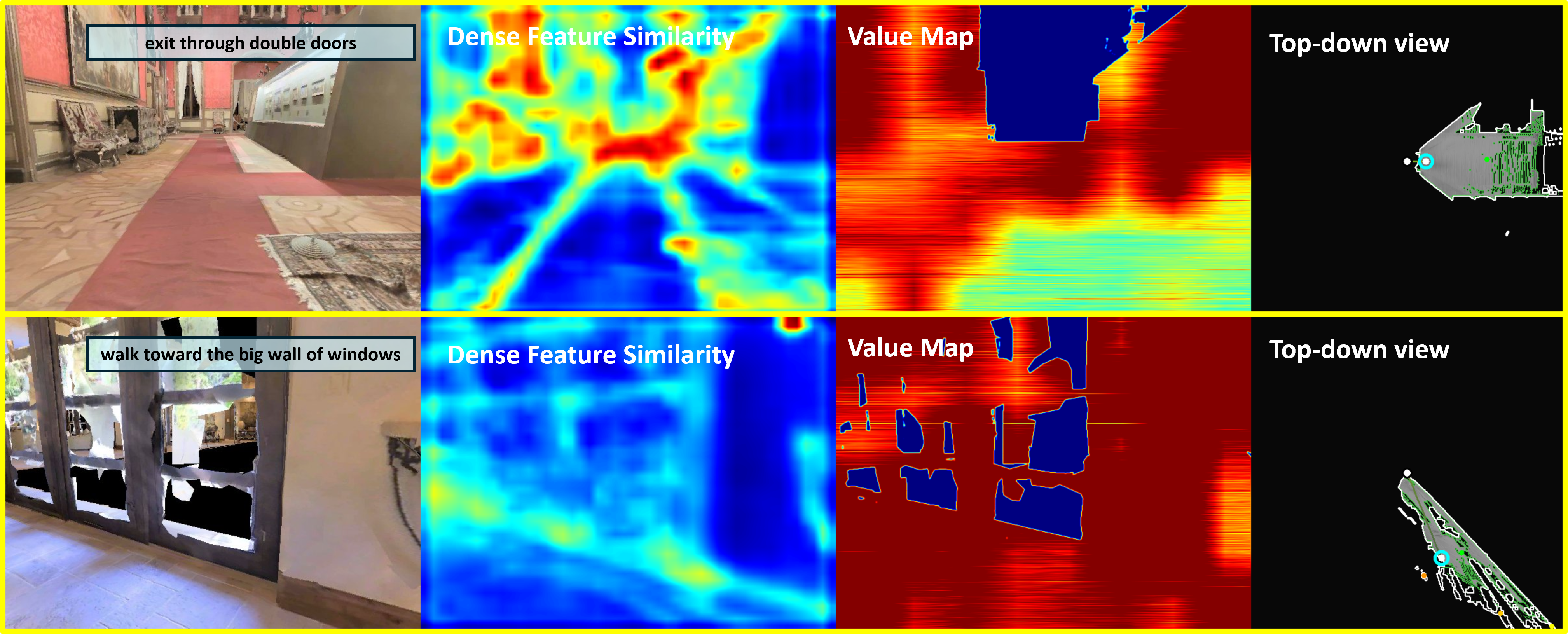}
\caption{Visualization of two representative navigation episodes at decision step 20. The active sub-instructions in the top and bottom rows are ``exit through double doors'' and ``walk toward the big wall of windows,'' respectively. From left to right, each row shows the current RGB observation and active sub-instruction, the pixel-wise CleanDIFT score map, the fused semantic value map, and the top-down navigation state derived from the online reconstruction map. The CleanDIFT and semantic value maps use the JET colormap, with warmer colors indicating higher responses or values. In the top-down view, gray regions denote traversable space, green boundaries denote walls or obstacles, white contours show SLIC superpixel boundaries, and colored dots mark the centroids of the five highest-valued superpixels. The historical trajectory changes gradually from blue at the starting position to green at the current position.}
\label{fig:NAVI}
\end{figure}

To examine how LG-VLN perceives the environment, estimates navigation values, and selects waypoints, we visualize intermediate results from two representative episodes at decision step 20, as shown in Fig.~\ref{fig:NAVI}. The two examples illustrate different stages of navigation. In the first, the target landmark has not yet entered the agent's view, whereas in the second, it is already visible. Comparing these cases shows how the navigation behavior shifts from exploration toward target-directed movement.

In Task 1, the active sub-instruction is ``exit through double doors,'' but the target doors are not yet visible in the current RGB observation. Because the target lies outside the field of view, the fused semantic value map does not contain a clear local high-value region. The system instead relies on accumulated map information and the surrounding geometry to continue frontier exploration. CleanDIFT still produces relatively strong responses around visually distinctive structures, such as carpets, furniture boundaries, and door frames, indicating that it captures local appearance patterns even when they are not directly related to the current target. In the top-down view, high-value waypoint candidates are mainly located near traversable frontiers of the reconstructed map. At this stage, the agent remains in exploration mode, and its planned motion follows the geometry of the observed environment.

In Task 2, the active sub-instruction is ``walk toward the big wall of windows,'' and the target landmark is already visible. With stronger correspondence between the instruction and the current observation, the fused semantic value map shows a clear high-value region around the windows, while regions farther from the target receive lower values. This distribution indicates that the value map can distinguish target-relevant regions once sufficient visual evidence is available. CleanDIFT also responds strongly to the window frames, boundaries, and nearby floor regions, providing local visual cues for semantic value fusion. In the top-down view, the highest-value waypoint candidates are concentrated in the direction of the windows and distributed across traversable space. The agent's trajectory follows the same direction, indicating that the fused visual-semantic representation provides useful guidance for local planning.

These two examples illustrate the complementary roles of CleanDIFT and the semantic value map. CleanDIFT provides local visual similarity cues from textures, boundaries, and other fine-grained structures. The semantic value map combines these cues with the language instruction, accumulated map information, and the current navigation state to produce a spatial value distribution for planning. When the target is not visible, accumulated map information supports continued exploration. Once the target enters the field of view, visual evidence and language semantics reinforce the corresponding regions, concentrating high values near the target and directing waypoint selection toward it. After projection onto the map and SLIC superpixel clustering, the selected high-value candidates remain within traversable regions and align with the agent's direction of motion. Together, these examples suggest that the proposed value fusion and waypoint generation procedure balances target relevance with navigational feasibility.

\subsection{Depth Estimation and 3D Reconstruction Results}
\label{sec:depth_eval}

\subsubsection{Depth Estimation Results}
\label{subsubsec:depth_metrics}

We use the metric depth provided by Habitat as ground truth to evaluate the dense depth estimates. Let $\hat d_t$ denote the raw depth prediction of VGGT at frame $t$, $\tilde d_t$ the corresponding metric depth after applying the global scale factor $s_0$ estimated during navigation initialization, and $d_t^{gt}$ the Habitat ground-truth depth map in meters. The metric depth used during navigation is given by

\begin{equation}
\tilde d_t = s_0 \hat d_t.
\end{equation}

Ground-truth depth is not available during navigation for frame-wise scale or offset correction. We therefore evaluate the depth estimates using the same fixed global scale $s_0$ used at deployment, comparing $\tilde d_t$ with $d_t^{gt}$. The resulting depth estimation errors are reported in Table~\ref{tab:depth_metrics}.

To distinguish errors from global scale estimation from those associated with local relative depth structure, Table~\ref{tab:depth_metrics_decomp} also reports $\mathrm{AbsRel}$ under two scale-alignment settings. $\mathrm{AbsRel}_{s_0}$ applies the global scale factor $s_0$ estimated during initialization to all frames, matching the deployment setting used for navigation. In contrast, $\mathrm{AbsRel}_{\mathrm{med}}$ aligns each frame independently before computing the error. Because this frame-wise alignment requires ground-truth depth, it is used only for post-hoc analysis.

Table~\ref{tab:depth_metrics_decomp} further reports the same metrics for pixels satisfying $d_{\mathrm{gt}} < 5\,\mathrm{m}$, providing a separate measure of depth estimation accuracy at the short ranges most relevant to navigation.

\begin{table}[htbp]
\centering
\caption{Depth estimation error over all valid observation frames. Metrics are
computed per frame and then aggregated across frames.}
\label{tab:depth_metrics}
\normalsize
\setlength{\tabcolsep}{6pt}
\renewcommand{\arraystretch}{1.12}

\begin{tabularx}{\linewidth}{l*{5}{C}}
\toprule
\textbf{Statistics}
& \textbf{AbsRel}\(\downarrow\)
& \textbf{RMSE (m)}\(\downarrow\)
& \textbf{MAE (m)}\(\downarrow\)
& \textbf{SILog}\(\downarrow\)
& \textbf{Pearson}\(\uparrow\) \\
\midrule
mean    & 0.5874 & 0.6339 & 0.5371 & 0.1486 & 0.9107 \\
median  & 0.2643 & 0.4863 & 0.3941 & 0.1128 & 0.9723 \\
std     & 1.0531 & 0.5314 & 0.4949 & 0.1176 & 0.1851 \\
25\%    & 0.1322 & 0.2696 & 0.2078 & 0.0691 & 0.9260 \\
75\%    & 0.4752 & 0.8248 & 0.6974 & 0.1890 & 0.9869 \\
\bottomrule
\end{tabularx}
\end{table}

\begin{table}[htbp]
\centering
\caption{Depth estimation errors under different scale alignment settings.}
\label{tab:depth_metrics_decomp}
\normalsize
\renewcommand{\arraystretch}{1.15}
\setlength{\tabcolsep}{4pt}

\begin{tabularx}{\linewidth}{*{5}{C}}
\toprule
\textbf{Statistics}
& \textbf{\(\mathrm{AbsRel}_{s_0}\)}
& \textbf{\(\mathrm{AbsRel}_{\mathrm{med}}\)}
& \textbf{\(\mathrm{AbsRel}_{s_0}^{<5\,\mathrm{m}}\)}
& \textbf{\(\mathrm{AbsRel}_{\mathrm{med}}^{<5\,\mathrm{m}}\)} \\
\midrule
mean    & 0.5874 & 0.0493 & 0.5875 & 0.0484 \\
median  & 0.2643 & 0.0319 & 0.2629 & 0.0299 \\
25\%    & 0.1322 & 0.0186 & 0.1302 & 0.0174 \\
75\%    & 0.4752 & 0.0586 & 0.4747 & 0.0558 \\
\bottomrule
\end{tabularx}
\end{table}

All metrics are computed over valid ground-truth pixels in each frame and then aggregated across frames. Table~\ref{tab:depth_metrics} therefore reflects frame-to-frame variability rather than a global pixel-level distribution. The fixed-scale $\mathrm{AbsRel}$ is strongly right-skewed, with a mean of $0.5874$ and a median of only $0.2643$, together with a large standard deviation of $1.0531$. This pattern suggests that the mean is driven by a relatively small number of frames with severe scale errors.

Table~\ref{tab:depth_metrics_decomp} further separates scale errors from errors in relative depth structure. After independent median alignment for each frame, $\mathrm{AbsRel}$ decreases from $0.5874$ to $0.0493$ on average and from $0.2643$ to $0.0319$ at the median. A similar reduction is observed for pixels with $d_{\mathrm{gt}}<5,\mathrm{m}$. The median SILog of $0.1128$ and Pearson correlation of $0.9723$ provide further evidence that most of the error comes from frame-to-frame scale variation rather than inaccurate relative depth structure.

The same pattern appears in Table~\ref{tab:geometry_ablation}. Replacing ground-truth depth with VGGT depth while retaining ground-truth poses results in only a small drop in SR. Approximately uniform changes in scale largely preserve bearings, depth ordering, and traversability structure, making navigation less sensitive to scale errors than metric reconstruction. Residual scale variation can still affect fixed metric thresholds, however, and the use of a single global scale remains a limitation.

\begin{figure}[htbp]
\centering
\includegraphics[width=\linewidth]{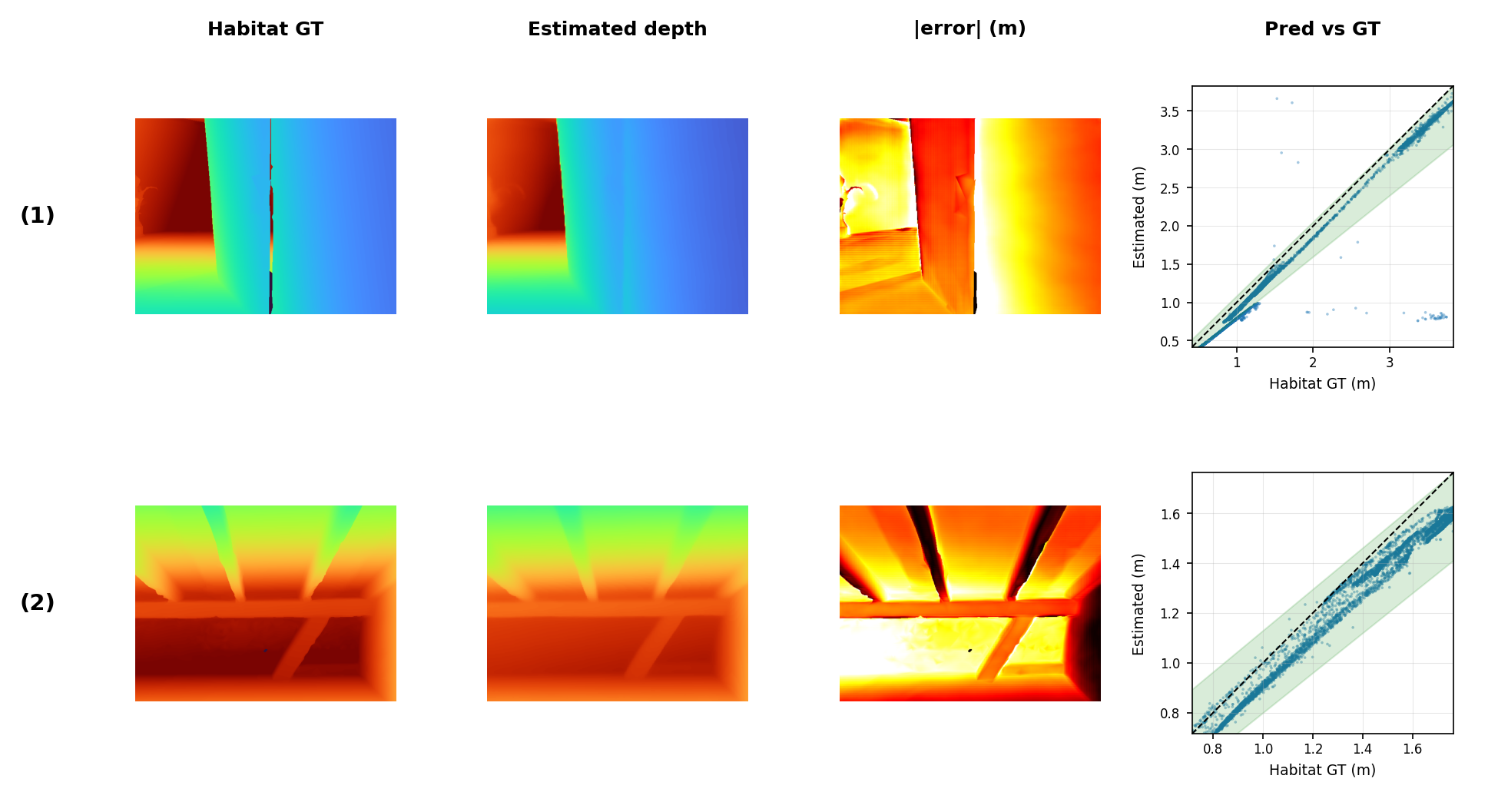}
\caption{Qualitative depth estimation results on representative R2R-CE frames. Each row corresponds to one observation frame and shows, from left to right: (a) the Habitat ground-truth depth map; (b) the estimated depth map after global metric-scale calibration; (c) the pixel-wise absolute error map; and (d) a scatter plot of the predicted depth against the ground-truth depth, including the dashed $y=x$ reference line and the tolerance band corresponding to $\delta < 1.25$.}
\label{fig:depth_examples}
\end{figure}

For qualitative evaluation, we select two representative R2R-CE navigation episodes and use the first valid observation frame from each episode. In Fig.~\ref{fig:depth_examples}, the first three columns show the Habitat ground-truth depth, the scale-corrected depth estimate, and the pixel-wise absolute error map. The fourth column plots the estimated depth against the ground-truth depth, together with the $y=x$ reference line and the tolerance region corresponding to $\delta < 1.25$. These two frames are near-range observations selected for visualization and are not representative of the full-split statistics reported in Table~\ref{tab:depth_metrics}.

Fig.~\ref{fig:depth_examples} shows that the estimated depth in both examples is strongly correlated with the ground-truth depth. However, the scatter plots remain systematically offset from the $y=x$ reference line, indicating that the fixed global scale does not fully match the effective scale of each frame. Large absolute errors occur mainly near object boundaries and depth discontinuities, where the predicted depth tends to be spatially smoother than the ground truth. 

Across all valid frames, the mean and median Pearson correlation coefficients are 0.9107 and 0.9723. The high correlation indicates that the predicted depth generally preserves relative depth variation and scene structure even when its absolute metric scale is inaccurate.

\subsubsection{3D Reconstruction Results}

\begin{figure}[htbp]
\centering
\includegraphics[width=\linewidth]{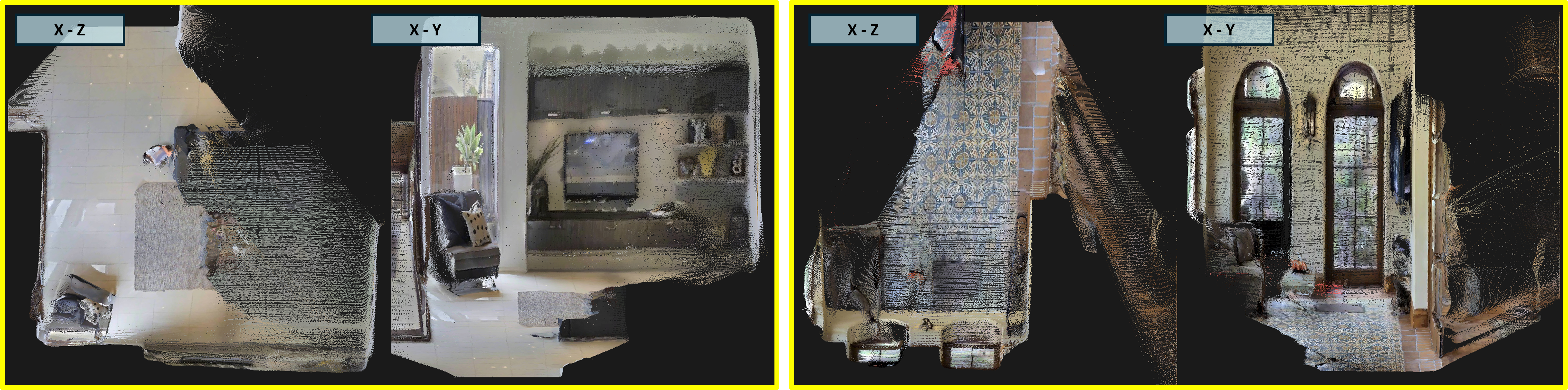}
\caption{3D reconstruction results for two representative indoor scenes with different spatial layouts. The X--Z panels show top-down views of the accumulated point clouds, highlighting the planar scene layout and spatial consistency. The X--Y panels show side views, illustrating scene height, depth variation, and local geometric structure.}
\label{fig:recon_3d}
\end{figure}

We qualitatively evaluate the 3D reconstructions obtained with VGGT on two representative navigation episodes with different spatial layouts, as shown in Fig.~\ref{fig:recon_3d}. In both cases, reconstruction uses only consecutive monocular RGB frames, without additional sensing modalities such as depth, IMU, or odometry. The accumulated point clouds are visualized from the X--Z top-down view and the X--Y side view to examine the scene layout, reconstructed geometry, and consistency across frames.

The example on the left shows an indoor room with a relatively regular layout. In the X--Y side view, the main structures, including the walls, floor, and large pieces of furniture, are largely recovered, with clear separation across different depth ranges. The X--Z top-down view preserves the room boundaries and the relative positions of the furniture, without obvious large-scale ghosting or structural misalignment.

The example on the right shows a more complex scene containing nearby furniture, a long corridor, and arched windows. The reconstruction preserves the main geometric contours of both the foreground furniture and the distant corridor while maintaining a largely continuous spatial layout. Some scattered points remain near wall boundaries and thin structures, but they do not introduce substantial geometric discontinuities.

Inter-frame consistency is maintained through $\mathrm{Sim}(3)$ alignment between adjacent local reconstructions and point-cloud filtering based on geometric confidence and CleanDIFT semantic consistency. The alignment compensates for differences in rotation, translation, and scale across local reconstructions, while the filtering step suppresses points arising from low-confidence predictions and incorrect matches. Low-texture regions and depth discontinuities can still produce locally sparse reconstructions or isolated outliers.

Overall, these examples show that VGGT can recover the main geometric structures of indoor scenes from consecutive monocular RGB images and integrate them into relatively continuous point clouds in a unified coordinate system. The reconstructed geometry provides a practical spatial representation for subsequent value-map construction and local path planning.

\subsection{Pose Estimation Accuracy Evaluation}
\label{sec:pose_eval}

The proposed system relies on online camera pose estimates for map updates and local path planning during navigation. In geometrically ambiguous scenes, incorrect correspondences can reduce pose estimation accuracy. To mitigate this problem, we use CleanDIFT features to filter correspondences based on semantic consistency. We quantitatively evaluate the resulting pose estimates using Absolute Trajectory Error (ATE) and the translational component of Relative Trajectory Error (RTE).

We evaluate the pose estimation pipeline under two configurations. In the w/o CleanDIFT setting, inter-frame $\mathrm{Sim}(3)$ registration relies only on geometric RANSAC. In the w/ CleanDIFT setting, CleanDIFT semantic consistency is used to filter the correspondences before geometric RANSAC. The $\mathrm{Sim}(3)$ transformation is then re-estimated from the remaining semantically consistent matches.

For evaluation, the estimated pose at each frame is compared with the corresponding ground-truth pose provided by Habitat. After $\mathrm{Sim}(3)$ alignment, ATE is computed as the root mean squared error (RMSE) between the estimated and ground-truth trajectories. RTE is computed from the relative translation error between consecutive frames and is also reported as RMSE. For each metric, we average the RMSE across all navigation episodes to obtain the overall result.

\begin{table}[h]
\centering
\caption{Pose estimation accuracy on the 550-episode evaluation subset.}
\label{tab:pose_metrics}
\small
\setlength{\tabcolsep}{12pt}
\renewcommand{\arraystretch}{1.12}

\begin{tabularx}{\linewidth}{*{3}{C}}
\toprule
\textbf{Mode}
& \textbf{ATE RMSE (m)}
& \textbf{RTE RMSE (m)} \\
\midrule
w/o CleanDIFT & 0.322 & 0.150 \\
w/ CleanDIFT  & 0.318 & 0.092 \\
\bottomrule
\end{tabularx}
\end{table}

Table~\ref{tab:pose_metrics} shows that the CleanDIFT semantic consistency constraint substantially improves relative pose estimation. The RTE RMSE decreases from \(0.150\,\mathrm{m}\) to \(0.092\,\mathrm{m}\),
corresponding to a \(38.7\%\) reduction, whereas the ATE RMSE is essentially
unchanged (\(0.322\,\mathrm{m}\) versus \(0.318\,\mathrm{m}\)). Since RTE more directly reflects local frame-to-frame motion accuracy, the larger improvement suggests that CleanDIFT mainly enhances inter-frame registration rather than global trajectory consistency.

In indoor scenes with repetitive structures or weak texture, geometric matching may produce ambiguous correspondences that are not fully rejected by RANSAC. CleanDIFT introduces semantic consistency as an additional criterion for correspondence filtering, reducing geometrically plausible but semantically incorrect matches. This improves the reliability of correspondences used for relative pose estimation and consequently reduces local translation errors. The limited improvement in ATE is expected because CleanDIFT operates primarily on local pairwise alignment rather than explicitly optimizing the complete trajectory.

\section{Conclusion}
\label{sec:conclusion}

This work addresses zero-shot VLN-CE using only a monocular RGB camera and presents LG-VLN, a framework that shares dense visual features across geometric estimation and semantic navigation. A feed-forward 3D reconstruction network estimates depth, camera poses, and dense point clouds online, which are used to update geometric and semantic maps in a unified world coordinate system. To connect geometric mapping with semantic navigation, CleanDIFT features are used in both stages. In the geometric front-end, semantic consistency filters inter-frame correspondences to reduce the effect of incorrect matches on pose estimation and map fusion. During navigation, target-conditioned visual reference similarity is combined with local image--text relevance from BLIP-2 to construct a semantic value map for the active sub-instruction.

LG-VLN also maintains an explicit shared state that records task progress, map states, perception results, and action feedback. LangGraph organizes instruction decomposition, constraint evaluation, semantic perception, incremental mapping, waypoint selection, path planning, action execution, and failure recovery into a directed workflow. Conditional routing governs sub-instruction transitions, termination decisions, and recovery procedures, allowing the navigation modules to operate through a common state and control interface.

On a fixed 550-episode subset of the R2R-CE val-unseen split, LG-VLN achieves an SR of 21.3\%, an SPL of 12.1\%, and an NE of $7.85\,\mathrm{m}$ without task-specific training data, ground-truth depth or poses, odometry, or prior maps. Ablation results show that pixel-level visual similarity from CleanDIFT and local image--text relevance from BLIP-2 provide complementary semantic signals. The same dense CleanDIFT features also support geometric correspondence filtering, allowing them to contribute to both geometric estimation and semantic value construction. The results further indicate that online pose estimation errors have a greater overall effect on navigation performance than depth prediction errors, with a particularly clear effect on NE. The difference in SR is relatively small under the current single-run evaluation, so a stronger conclusion would require more episodes or repeated runs. Overall, these results show that zero-shot VLN-CE can be performed from monocular RGB observations alone by combining online geometric estimation, shared visual features, and semantic value fusion.

\appendix
\numberwithin{equation}{section}

\section{Implementation Details}
\label{app:implementation}

This appendix provides implementation details that complement the method description in the main paper. We focus on coordinate conventions, event-driven perception scheduling, monocular scale initialization, constraint evaluation, semantic-value construction, map fusion, and the recovery policy. Routine software-level control flow is omitted unless it affects the behavior or reproducibility of the proposed method.

\subsection{Coordinate Systems and Map Projection}
\label{app:coordinate_transform}
We use $\mathcal T_a^b$ to denote the transformation from coordinate frame $a$ to coordinate frame $b$. Because the world-aligned map is maintained in a similarity-coordinate representation, transformations between the reconstructed camera frames and the world frame are represented in $\mathrm{Sim}(3)$. A similarity transformation is written as
\begin{equation}
\mathcal T_a^b
=
\left(
s_a^b,
\mathbf R_a^b,
\mathbf t_a^b
\right)
\in
\mathrm{Sim}(3),
\end{equation}
and acts on a 3D point according to
\begin{equation}
{}^b\mathbf x
=
s_a^b
\mathbf R_a^b
{}^a\mathbf x
+
\mathbf t_a^b .
\label{eq:sim3_action_app}
\end{equation}
Here, $s_a^b>0$ denotes the scale factor, $\mathbf R_a^b\in\mathrm{SO}(3)$ denotes the rotational component, and $\mathbf t_a^b\in\mathbb R^3$ denotes the translation expressed in frame $b$. Standard composition and inversion of similarity transformations are used throughout the system.

At decision step $t$, the agent state in the world-aligned map is represented by the body-to-world similarity transformation
$\mathcal T_{\mathcal B_t}^{\mathcal W}\in\mathrm{Sim}(3)$,
written in homogeneous form as
\begin{equation}
\mathcal T_{\mathcal B_t}^{\mathcal W}
=
\begin{bmatrix}
s_t^{\mathcal W}\mathbf R_t^{\mathcal W} & \mathbf p_t^{\mathcal W} \\
\mathbf 0^\top & 1 
\end{bmatrix} \in \mathrm{Sim}(3), \label{eq:agent_pose_app} 
\end{equation} 
where $s_t^{\mathcal W}>0$ denotes the scale associated with the current world-map representation, $\mathbf R_t^{\mathcal W}\in\mathrm{SO}(3)$ denotes the orientation of the body frame $\mathcal B_t$ with respect to the world frame $\mathcal W$, and $\mathbf p_t^{\mathcal W}=(x_t,y_t,z_t)^\top\in\mathbb R^3$ denotes the origin of the body frame expressed in the world frame. 

The scale component $s_t^{\mathcal W}$ is a property of the similarity-based map representation rather than a physical degree of freedom of the agent. It accounts for residual scale variations between monocular reconstructions while preserving the rotational and translational components required for navigation. The body and camera frames are related through a fixed camera-to-body extrinsic transformation $\mathcal T_{\mathcal C}^{\mathcal B}$. The extrinsic transformation represents a rigid calibration and therefore belongs to $\mathrm{SE}(3)$, which is naturally embedded in $\mathrm{Sim}(3)$ with unit scale. The camera-to-world transformation is consequently obtained by 
\begin{equation} 
\mathcal T_{\mathcal C_t}^{\mathcal W} = \mathcal T_{\mathcal B_t}^{\mathcal W} \circ \mathcal T_{\mathcal C}^{\mathcal B}, \qquad \mathcal T_{\mathcal C_t}^{\mathcal W} \in \mathrm{Sim}(3), \label{eq:camera_pose_app} 
\end{equation} 
with the inverse world-to-camera transformation given by 
\begin{equation} 
\mathcal T_{\mathcal W}^{\mathcal C_t} = \left( \mathcal T_{\mathcal C_t}^{\mathcal W} \right)^{-1}. \label{eq:camera_pose_inverse_app} 
\end{equation} 
Here, $\mathcal T_{\mathcal C}^{\mathcal B}\in\mathrm{SE}(3)$ is fixed throughout an episode and does not introduce any additional scale variation. For a pixel $p=(u,v)$ with homogeneous coordinate $\bar{\mathbf p}=[u,v,1]^\top$, the corresponding reconstructed point in the camera frame is obtained from its predicted depth as 
\begin{equation} 
\mathbf x_t^{\mathcal C}(p) = Z_t(p)\mathbf K_{\rm cam}^{-1}\bar{\mathbf p}, \label{eq:pixel_camera_point_app} 
\end{equation} 
where $Z_t(p)$ denotes the reconstructed depth and $\mathbf K_{\rm cam}$ is the camera intrinsic matrix. The corresponding point in the world-aligned map is then obtained through the similarity transformation 
\begin{equation} 
\mathbf x_t^{\mathcal W}(p) = \mathcal T_{\mathcal C_t}^{\mathcal W} \left( \mathbf x_t^{\mathcal C}(p) \right). \label{eq:pixel_to_world_app} 
\end{equation} 
Thus, the same world-aligned similarity representation is used for accumulating reconstructed geometry and for projecting semantic observations into the global map. For ground-plane mapping, we use the horizontal world coordinates $(x,z)$. The horizontal component of the agent position is therefore 
\begin{equation} 
\mathbf p_{t,\mathrm{h}}^{\mathcal W} = (x_t,z_t)^\top. \label{eq:horizontal_agent_position_app} 
\end{equation} 
A point on the ground plane is represented by $\mathbf q^{\mathcal W}=(x,z)^\top$. Because the local map remains aligned with the horizontal axes of the world frame, its coordinates are obtained by translating the world coordinates by the current horizontal agent position: 
\begin{equation} 
\mathbf q^{\mathcal L_t} = \mathbf q^{\mathcal W} - \mathbf p_{t,\mathrm{h}}^{\mathcal W} = (x-x_t,\;z-z_t)^\top. \label{eq:local_map_transform_app} 
\end{equation} 
The local-map orientation is therefore fixed to the world frame, while its origin follows the agent. Agent rotation does not rotate the local map. For a map resolution of $r$ meters per cell and local-map center $\mathbf o=(o_r,o_c)$, a horizontal world coordinate is projected to the local grid as 
\begin{equation} 
\pi_t(\mathbf q^{\mathcal W}) = \left( \operatorname{round}\!\left(\frac{z-z_t}{r}\right)+o_r,\; \operatorname{round}\!\left(\frac{x-x_t}{r}\right)+o_c \right), \label{eq:gtm_project_app} 
\end{equation} 
with the corresponding inverse mapping from a grid-cell center given by 
\begin{equation} 
\pi_t^{-1}(i,j) = \left( x_t+(j-o_c)r,\; z_t+(i-o_r)r \right)^\top . \label{eq:gtm_unproject_app} 
\end{equation} 
For temporal reprojection, the transformation from the previous camera frame $\mathcal C_{t-1}$ to the current camera frame $\mathcal C_t$ is obtained from the corresponding world-frame similarity transformations: 
\begin{equation} 
\mathcal T_{\mathcal C_{t-1}}^{\mathcal C_t} = \left( \mathcal T_{\mathcal C_t}^{\mathcal W} \right)^{-1} \circ \mathcal T_{\mathcal C_{t-1}}^{\mathcal W} \in \mathrm{Sim}(3). \label{eq:relative_sim3_app} 
\end{equation} 
This relative transformation captures the rotation, translation, and residual scale variation between consecutive monocular reconstructions. It is used consistently for inter-frame geometric registration, geometric consistency verification, and reprojection of historical semantic observations into the current camera view.

\subsection{Perception Scheduling and Navigation Failure Detection}
\label{app:perception_failure}

\paragraph{Event-driven semantic perception.}
Geometric reconstruction is updated for every valid RGB observation, whereas open-vocabulary semantic perception is invoked only when it is expected to provide substantially new information. Let $t_{\rm sem}$ denote the time of the most recent semantic update. A new semantic observation is requested at episode initialization, immediately after a sub-instruction transition, after recovery, or whenever the cached semantic observation does not correspond to the active sub-instruction.

Semantic perception is also refreshed when the viewpoint has changed sufficiently since $t_{\rm sem}$. Specifically, we trigger an update after a translation of at least $\tau_p$ or an accumulated heading change of at least $\tau_\psi$. An additional update is requested when the current constraint decision is ambiguous, for example, when an object detection score is close to its acceptance threshold or when a location constraint has not accumulated sufficient positive or negative evidence.

A semantic cache is therefore considered valid only if it was generated for the currently active sub-instruction. This prevents object categories, visual references, or relevance estimates associated with the previous navigation stage from influencing the current one.

The viewpoint thresholds are chosen to maintain sufficient spatial coverage. If $\phi_h$ denotes the horizontal field of view of the camera, we use
$\tau_\psi<\phi_h$.
Likewise, if the effective perception range is lower-bounded by $d_{\min}$, the translational spacing is chosen such that
\begin{equation}
\tau_p
<
d_{\min}\tan\!\left(\frac{\phi_h}{2}\right).
\label{eq:perception_coverage_app}
\end{equation}
In our experiments, $\tau_\psi=30^\circ$ and $\tau_p=0.30\,\mathrm m$.

\paragraph{Motion consistency.}
After each action, the expected displacement is compared with the motion estimated from visual registration. Forward actions that repeatedly produce substantially less translation than the commanded step provide evidence that the corresponding local free-space estimate may be incorrect. Cells immediately along the attempted motion direction are therefore assigned additional obstacle evidence during map fusion.

The same signal also serves as an indicator of navigation stagnation. We do not declare a failure from a single inconsistent action because occasional reconstruction errors, collisions, or imperfect correspondence estimates can occur during normal navigation.

\paragraph{Persistent failure detection.}
The graph monitors four complementary indicators over short temporal windows: ineffective forward motion, spatial recurrence, heading oscillation, and lack of waypoint progress. Spatial recurrence is detected when the current pose returns to the neighborhood of a recently visited pose. Heading oscillation captures repeated alternating or large heading changes that do not produce meaningful translational progress. Initial panoramic scans and explicit direction-changing constraints are excluded from this test.

Waypoint progress is evaluated only while the high-level target remains unchanged. Let
$d_t^{\rm geo}$ denote the geodesic distance from the current position to the active waypoint. The agent is considered to make progress if this distance decreases sufficiently over the monitoring window or if progress is made toward satisfying the active instruction constraints during the same interval. When the waypoint changes, the progress window is restarted for the new target.

A small stagnation counter aggregates these signals over time. Negative events increase the counter, whereas sustained effective motion and waypoint progress decrease it. Recovery is invoked only when the counter exceeds the stagnation threshold, avoiding unnecessary interruptions caused by isolated failures.

\subsection{Monocular Scale Recovery and Registration}
\label{app:slam_details}

\paragraph{Ground-based metric initialization.}
VGGT produces geometry with an initially unknown metric scale. We recover this scale from the known camera mounting height $h_{\rm cam}$ using ground observations collected during initialization.

For each initialization view, ground pixels with sufficient VGGT confidence are back-projected into the camera frame:
\begin{equation}
\mathcal G_t
=
\left\{
\hat Z_t(p)\mathbf K_{\rm cam}^{-1}\bar{\mathbf p}
\;\middle|\;
\mathbf M_t^{\rm gnd}(p)=1,\;
C_t^{\rm vggt}(p)>\tau_{\rm geo}
\right\}.
\label{eq:ground_points_app}
\end{equation}
RANSAC is first used to remove non-ground points. A plane
$\widehat{\mathbf n}_t^\top\mathbf x+\widehat b_t=0$
is then fitted to the remaining points, with VGGT geometric confidence used as the fitting weight.

For a normalized plane normal, $|\widehat b_t|$ gives the reconstructed camera-to-ground distance. Each valid frame therefore yields a scale estimate
\begin{equation}
\widehat{s}_t
=
\frac{h_{\mathrm{cam}}}{\left|\widehat{b}_t\right|},
\qquad
s_0
=
\operatorname{median}
\left(
\left\{
\widehat{s}_t
\mid
t\in\mathcal F_{\mathrm{valid}}
\right\}
\right).
\label{eq:initial_scale_app}
\end{equation}

A scale observation is accepted only if the ground plane contains at least
$N_{\min}^{\rm gnd}$ inliers, its normal is sufficiently close to the expected vertical direction, the weighted plane-fitting residual is below the prescribed tolerance, and the resulting scale lies within the admissible interval $[s_{\min},s_{\max}]$.

Scale observations are collected throughout the initial panoramic scan. The nominal initialization uses $T_{\rm init}=12$ headings. If fewer than three reliable scale estimates are available, additional observations are collected until at least three valid measurements have been obtained. Their median $s_0$ is then used as the episode-level metric scale.

\subsubsection{CleanDIFT-based Semantic Consistency Filtering}
\label{subsec:app_cleandift_details}

For each candidate correspondence in $\mathcal C_t^{0}$ generated by the geometric matching module between consecutive frames, the system measures semantic consistency using the CleanDIFT descriptors at the corresponding image locations. The semantic similarity is computed as
\begin{equation}
\gamma_i
=
\frac{
\mathbf f_{t-1}
\left(
\mathbf u_i^{t-1}
\right)^{\mathsf T}
\mathbf f_t
\left(
\mathbf u_i^{t}
\right)
}{
\left\|
\mathbf f_{t-1}
\left(
\mathbf u_i^{t-1}
\right)
\right\|_2
\left\|
\mathbf f_t
\left(
\mathbf u_i^{t}
\right)
\right\|_2
}.
\label{eq:app_semantic_similarity}
\end{equation}
Since the descriptors are $\ell_2$-normalized, $\gamma_i$ is equivalent to their inner product.

The semantically and geometrically consistent correspondence set is defined as
\begin{equation}
\mathcal C_t^{\mathrm{sem}}
=
\left\{
c_i\in\mathcal C_t^{0}
\;\middle|\;
\gamma_i\geq\tau_{\mathrm{sem}},\;
C_i^{t-1}\geq\tau_{\mathrm{geo}},\;
C_i^{t}\geq\tau_{\mathrm{geo}}
\right\},
\label{eq:app_semantic_filtering}
\end{equation}
where $\tau_{\mathrm{sem}}$ and $\tau_{\mathrm{geo}}$ denote the semantic similarity and geometric-confidence thresholds, respectively, and $C_i^{t-1}$ and $C_i^{t}$ are the VGGT geometric confidence values of the two corresponding 3D points.

This filtering combines semantic consistency from CleanDIFT with geometric reliability from VGGT. The semantic criterion removes correspondences that are visually inconsistent despite geometric agreement, while geometric confidence suppresses points with unreliable reconstruction quality. The semantic threshold is selected such that $\tau_{\mathrm{sem}}>0$, ensuring that the similarity values used in subsequent correspondence weighting remain non-negative.

\paragraph{Similarity alignment.}
Given the filtered correspondence set
$\mathcal C_t^{\mathrm{sem}}$, RANSAC first provides a robust initial estimate
of the inter-frame similarity transformation while removing geometrically
inconsistent correspondences. The transformation is then refined by solving a
confidence-weighted robust optimization problem:
\begin{equation}
\left(
\hat s_{t-1,t},
\hat{\mathbf R}_{t-1,t},
\hat{\mathbf t}_{t-1,t}
\right)
=
\underset{
s_{t-1,t},
\mathbf R_{t-1,t},
\mathbf t_{t-1,t}
}{
\operatorname{argmin}
}
\sum_{c_i\in\mathcal C_t^{\mathrm{sem}}}
w_i
\rho
\left(
\left\|
\mathbf x_i^{t}
-
\left(
s_{t-1,t}
\mathbf R_{t-1,t}
\mathbf x_i^{t-1}
+
\mathbf t_{t-1,t}
\right)
\right\|_2^2
\right),
\label{eq:app_robust_sim3}
\end{equation}
where $\hat s_{t-1,t}$, $\hat{\mathbf R}_{t-1,t}$, and
$\hat{\mathbf t}_{t-1,t}$ denote the estimated scale, rotation, and translation,
respectively, and $\rho(\cdot)$ represents the robust loss function.

Each correspondence is weighted according to both geometric confidence and
semantic consistency:
\begin{equation}
w_i
=
\min
\left(
C_i^{t-1},
C_i^{t}
\right)
\cdot
\gamma_i,
\label{eq:app_matching_weight}
\end{equation}
where $w_i$ measures the contribution of correspondence $c_i$ to the robust
optimization. Taking the minimum of the two geometric confidence values
ensures that a correspondence receives a lower weight when either endpoint is
unreliable, while $\gamma_i$ reduces the influence of semantically inconsistent
matches.

The estimated transformation is accepted only if the RANSAC consensus set
contains at least $N_{\min}^{\mathrm{inlier}}$ valid correspondences and the
estimated scale satisfies the bounded scale-variation constraint:
\begin{equation}
\left|
\log \hat s_{t-1,t}
\right|
\leq
\epsilon_s.
\label{eq:app_scale_constraint}
\end{equation}

If RANSAC fails or the number of valid inliers is below
$N_{\min}^{\mathrm{inlier}}$, the previous pose estimate is retained, and the
current geometric observation is excluded from fusion into the global metric
map.

\subsection{Constraint Evaluation Details}
\label{app:constraint_details}

The constraint manager uses different temporal evidence models for object,
location, and direction constraints. Instead of switching a sub-instruction
based on a single semantic prediction, it accumulates evidence across
observations and combines it with the estimated trajectory.

\paragraph{Object constraints.}
Grounding DINO provides open-vocabulary bounding boxes, and SAM2 refines each
accepted detection into an instance mask. Mask pixels with valid reconstructed
geometry are transformed into the world frame, and the target position is
robustly estimated from the resulting 3D points.

To reduce isolated false detections, an object constraint is confirmed only
after consistent support is obtained across multiple semantic perception
events. The target must also satisfy the spatial relationship specified by the
instruction. For ordinary approach-type constraints, the agent must be within
the effective target range $r_{\rm obj}$ before the constraint is marked as
complete. The target position is updated whenever additional reliable
observations become available.

\paragraph{Location constraints.}
For each semantic region $\ell$, we maintain a compact state
\begin{equation}
q_\ell^t
\in
\{
\mathrm{unknown},
\mathrm{inside},
\mathrm{outside}
\}.
\label{eq:location_state_app}
\end{equation}
BLIP-based VQA provides the primary region-membership observation, while
BLIP-2 image--text relevance is used when direct VQA evidence is unavailable
or unreliable.

Raw predictions are temporally filtered. A transition into the
\emph{inside} state requires $\kappa_{\rm in}$ consecutive positive
observations, whereas a transition from \emph{inside} to \emph{outside}
requires $\kappa_{\rm out}$ consecutive negative observations. The system also
records whether the region has been previously visited and whether a stable
entry has occurred during the current sub-instruction.

This compact state representation supports the location expressions required by
VLN instructions. An entry constraint is satisfied by a stable transition into
the region; an exit constraint requires a stable transition from inside to
outside; and a visited constraint remains satisfied once a stable visit has
occurred. A traversal constraint is stricter: the agent must enter the region
during the current sub-instruction, remain inside it for the required duration,
and subsequently exit. Therefore, a single positive VQA prediction cannot
satisfy a traversal instruction.

When a new sub-instruction becomes active, long-term region visitation
information is preserved, while sub-instruction-specific entry and duration
records are reset.

\paragraph{Turn constraints.}
Explicit turning commands are evaluated from heading changes. When
sub-instruction $k$ starts, the current heading $\psi_k^{\rm ref}$ is stored.
A turn constraint is satisfied only when the accumulated signed heading change
relative to this reference falls within the angular interval associated with
the requested turn. These constraints depend only on orientation and do not
require translational motion.

\paragraph{Relative motion constraints.}
Instructions such as ``move left'' or ``continue forward'' describe
displacement relative to the heading at the beginning of the sub-instruction.
Let $\mathbf v_{{\rm ref},k}$ denote this reference heading on the ground plane.
Over a temporal window of length $\tau$, we compute
\begin{equation}
\mathbf v_{\rm disp}^t
=
\mathbf p_t^{\mathcal W}
-
\mathbf p_{t-\tau}^{\mathcal W},
\qquad
\theta_t^{\rm mot}
=
\arccos
\frac{
\langle
\mathbf v_{{\rm ref},k},
\mathbf v_{\rm disp}^t
\rangle
}{
\|\mathbf v_{{\rm ref},k}\|_2
\|\mathbf v_{\rm disp}^t\|_2
}.
\label{eq:motion_angle_app}
\end{equation}
Direction is evaluated only when
$\|\mathbf v_{\rm disp}^t\|_2\ge\epsilon_{\rm disp}$.
The side of the reference direction is determined from
\begin{equation}
\varsigma_t
=
v_{{\rm ref},k,x}\Delta z_t
-
v_{{\rm ref},k,z}\Delta x_t .
\label{eq:motion_side_app}
\end{equation}

We classify motion using angular boundaries of
$15^\circ$ and $120^\circ$. Displacements with
$\theta_t^{\rm mot}<15^\circ$ are classified as forward, whereas
$\theta_t^{\rm mot}\ge120^\circ$ are classified as backward.
Intermediate directions are classified as left or right according to the sign
of $\varsigma_t$.

This displacement-based definition is intentionally separated from turn
evaluation. An in-place rotation therefore cannot satisfy a relative motion
constraint, and lateral displacement cannot satisfy an explicit turn
constraint. Backward instructions are normalized by the instruction parser
into a turn followed by forward motion, while the backward category above is
retained only for interpreting observed displacement.

\subsection{Construction of the Semantic Value Map}
\label{app:value_details}

\paragraph{Source normalization.}
The CleanDIFT relevance map and BLIP-2 image--text relevance map are
independently normalized to $[0,1]$ over geometrically valid pixels before
fusion in Eq.~\eqref{eq:semantic_evidence_fusion}. Pixels removed by geometric
confidence, visibility, or traversability filtering are excluded from the
normalization process.

When the valid value range of a relevance map is smaller than $\delta_x$, the
map is considered insufficiently discriminative for spatial guidance, and the
corresponding source reliability is set to zero. This prevents nearly uniform
similarity responses from dominating semantic-value estimation simply because
their absolute scores are high.

\paragraph{Visual-reference relevance.}
CleanDIFT visual-reference maps are constructed on the native feature grid.
Instance masks associated with the active object or landmark are resized to
this grid and used to aggregate reference descriptors. Dense cosine similarity
with the resulting reference descriptor produces $\mathbf A_t$, which is then
resized to the navigation-image resolution using bilinear interpolation.

For BLIP-2, the current observation is evaluated against the language
representation of the active constraint. Local image regions are scored
independently and interpolated to obtain the dense relevance map $\mathbf B_t$.
The source reliability terms $q_t^a$ and $q_t^b$ in
Eq.~\eqref{eq:semantic_evidence_fusion} are set to zero whenever the
corresponding source is unavailable or fails the validity checks described
above.

\paragraph{Observation confidence.}
The observation-confidence map integrates geometric validity, pose reliability,
viewing direction, observation distance, surface consistency, and
traversability. Pixels near the optical axis and those supported by reliable
reconstruction and registration receive higher weights, whereas uncertain
geometry, grazing-angle observations, obstacle regions, and unreliable
reprojections are suppressed.

Historical image-space semantic values are transferred to the current frame
using Eq.~\eqref{eq:relative_sim3_app}. A reprojected sample is retained only
if the transformed 3D point has positive depth, lies within the current image
boundary, and is consistent with the current reconstructed depth under the
occlusion-consistency tolerance. Samples violating any of these conditions are
removed by $\mathbf M_t^{\rm rep}$ before the confidence-weighted temporal
update in Eq.~\eqref{eq:image_value_temporal}.

This reprojection is used to stabilize semantic evidence across nearby
viewpoints rather than to maintain an indefinitely persistent image-space
memory. Long-term semantic information is instead stored in the
world-aligned global map.

\subsection{Map Fusion and Candidate Waypoint Generation}
\label{app:map_planning_details}

\subsubsection{Local Window and Global Fusion}

The fixed-size local map is centered on the current agent position and remains
aligned with the world axes. As the agent moves, the local window is translated
according to Appendix~\ref{app:coordinate_transform}. Cells entering the window
are recovered from the extensible global map whenever previous observations are
available; otherwise, they are initialized as unknown.

Traversability and occupancy evidence are accumulated in separate layers.
Occupancy is represented using a temporally decayed log-odds update:
\begin{equation}
\ell_t(\mathbf x)
=
\lambda_{\rm occ}\,
\ell_{t-1}(\mathbf x)
+
\ell_{\rm obs}(\mathbf x),
\label{eq:log_odds_update}
\end{equation}
where $0<\lambda_{\rm occ}\le1$. Occupied observations contribute positive
log-odds evidence, while free-space support is maintained independently in the
traversability layer. The occupancy probability is recovered through the
logistic transform.

This separation allows outdated obstacle evidence to decay while preventing
uncertain free-space observations from immediately overriding previously
observed obstacles. After independent fusion, occupied cells mask the
traversability layer, ensuring that occupancy takes precedence when the two
sources disagree.

The action--motion consistency signal provides an additional geometric
correction mechanism. When repeated forward commands produce negligible
visually estimated displacement, cells immediately along the attempted path
receive additional occupancy evidence. This correction only modifies the local
geometric interpretation of the failed motion and does not affect semantic
values.

Semantic values are fused into the same world-aligned cells using observation
confidence. Dense semantic features are averaged across repeated observations
and re-normalized after fusion. Observation count and exploration state are
maintained independently, allowing semantic confidence to be distinguished
from geometric coverage.

When a sub-instruction changes, target-specific semantic values from the
previous stage are attenuated rather than removed. The geometric map and
exploration history remain unchanged.

\subsubsection{Semantic-Enhanced Candidate Generation}

Waypoint candidates are generated only within the safe reachable region
$\mathcal R_t$ defined in Eq.~\eqref{eq:feasible_region}. Instead of selecting
isolated local maxima, we partition the reachable semantic map into spatially
coherent regions.

The clustering distance combines spatial proximity, semantic value, and dense
semantic-feature similarity:
\begin{equation}
d_{\rm SLIC}
=
d_{\rm spatial}
+
\lambda_1 d_{\rm value}
+
\lambda_2 d_{\rm sem},
\qquad
d_{\rm sem}
=
1-
\left\langle
\mathbf F_t^{\rm map}(\mathbf x),
\mathbf F_t^{\rm map}(\mathbf x')
\right\rangle .
\label{eq:semantic_slic_app}
\end{equation}
Each cluster is intersected with $\mathcal R_t$ and further divided into
8-connected components. Components with areas smaller than $A_{\min}$ are
discarded. For each remaining component, the navigable cell closest to its
geometric center is selected as the representative waypoint.

This procedure separates candidate construction from candidate scoring. The
semantic score used by the high-level planner remains the region-level score
defined in Eq.~\eqref{eq:new_waypoint_best}.

\subsubsection{Waypoint History and Loop Suppression}

The planner maintains recent pairs of agent positions and selected waypoints in
world coordinates. A newly generated waypoint is considered a loop candidate
when the agent returns to the neighborhood of a previous decision state and the
candidate is also close to the waypoint selected at that state. The
corresponding distance tolerances are $d_{\rm cyc,pos}$ and $d_{\rm cyc,wp}$.

Loop candidates are suppressed only during waypoint selection; the semantic and
geometric maps remain unchanged. If history filtering removes all otherwise
valid candidates, the planner falls back to the highest-scoring safe candidate
before loop suppression. This prevents the history mechanism from introducing
an artificial dead end.

Waypoint hysteresis is applied after candidate scoring. The previous waypoint
is retained as long as it remains reachable, unreached, and semantically
competitive with newly generated candidates. It is replaced only when it
becomes unsafe, is reached, loses sufficient semantic support, or is associated
with a detected navigation failure.

\subsection{LangGraph Execution and Recovery}
\label{app:recovery_details}

LangGraph implements the stateful execution policy described in
Sec.~\ref{sec:graph_framework}. The graph maintains persistent episode state
instead of treating perception, planning, and recovery as independent
operations.

\paragraph{Persistent and transient state.}
Persistent state includes the active sub-instruction index, estimated camera
and body poses, the global semantic-geometric map, constraint states,
exploration history, and waypoint history. These variables are preserved across
transitions between normal navigation and recovery.

Transient state contains the currently selected waypoint, immediate low-level
plans, perception-cache status, and recovery-control variables. These
quantities can be invalidated and recomputed without discarding the spatial
knowledge accumulated earlier in the episode.

\paragraph{Recovery policy.}
Recovery follows the three levels described in
Sec.~\ref{subsec:recovery_policy}. The levels are executed progressively rather
than simultaneously.

\begin{table}[htbp]
\centering
\caption{Summary of the recovery policy.}
\label{tab:recovery_policy}
\small
\begin{tabularx}{\linewidth}{@{}lXX@{}}
\toprule
Stage & Revision & Return condition \\
\midrule
$\mathrm{R}_1$: Viewpoint recovery
&
Perform a panoramic observation at the current pose and recompute
instruction-relevant visual evidence from the newly observed headings.
&
Reliable task-relevant evidence provides a valid reachable waypoint.
\\

$\mathrm{R}_2$: Semantic recovery
&
Invalidate semantic observations associated with the current target hypothesis
and reconstruct grounding, visual references, and relevance estimates from
recent observations.
&
A refreshed semantic hypothesis provides a valid reachable waypoint.
\\

$\mathrm{R}_3$: Navigation redirection
&
Temporarily select a reachable frontier or unexplored location within the safe
connected region while preserving the active instruction.
&
The intermediate target provides a new viewpoint that allows normal semantic
planning to resume.
\\
\bottomrule
\end{tabularx}
\end{table}

Viewpoint recovery is attempted first because it preserves all persistent
semantic and geometric states. If it does not recover sufficiently reliable
target evidence, semantic recovery is invoked. Navigation redirection is used
only when the current semantic hypothesis still cannot produce a useful
reachable waypoint.

The frontier selected in $\mathrm R_3$ serves as an intermediate exploration
target rather than a replacement for the instruction goal. Reaching this target
therefore triggers perception refresh and resumes the interrupted
sub-instruction.

After successful recovery, the global map, accumulated semantic memory,
constraint history, and waypoint history are preserved. The temporary
waypoint, pending low-level action, and stagnation state are cleared. Semantic
perception is then forced to refresh before returning to the normal navigation
loop. Geometry, map projection, and local planning are recomputed from the
updated state when necessary.

Each recovery level is attempted conservatively to avoid repeatedly applying
the same ineffective correction. If recovery remains unsuccessful after
$R_{\max}$ attempts, the episode is marked as unrecoverable and terminated. In
our experiments, $R_{\max}=3$.

\section{Model and Hyperparameter Configuration}
\label{app:hyperparams}

We evaluate the proposed system in the zero-shot setting on the 550-episode subset of the R2R-CE val-unseen split described in Sec.~\ref{subsec:simulation}. All perception and foundation-model components use publicly
available pretrained weights, and no R2R-CE navigation trajectories are used
for task-specific training.

Table~\ref{tab:hyperparams} summarizes the main thresholds and navigation
parameters required to reproduce the proposed method. Parameters used only for
numerical stability are set to standard implementation defaults and omitted
unless they influence decision boundaries.

\begin{table}[htbp]
\centering
\caption{Core model and hyperparameter settings.}
\label{tab:hyperparams}
\small
\begin{tabularx}{\textwidth}{@{}l c X@{}}
\toprule
Parameter & Value & Description \\
\midrule

\multicolumn{3}{c}{\textit{Geometric Reconstruction and Registration}} \\
\midrule
CleanDIFT similarity threshold $\tau_{\rm sem}$
&
$0.65$
&
Minimum cosine similarity required to retain semantically consistent
inter-frame correspondences.
\\

VGGT confidence threshold $\tau_{\rm geo}$
&
$0.10$
&
Minimum geometric confidence used for ground-plane estimation and
correspondence filtering.
\\

Point-cloud confidence percentile
&
$25\%$
&
Points within the lowest confidence quartile are removed before
high-confidence geometric processing.
\\

RANSAC inlier threshold
&
$0.10\ \mathrm m$
&
Metric residual threshold used for robust inter-frame geometric verification.
\\

Initialization scan directions $T_{\rm init}$
&
$12$
&
Uniformly spaced headings used during the initial panoramic scan and scale
initialization.
\\

\midrule
\multicolumn{3}{c}{\textit{Semantic Perception and Constraint Evaluation}} \\
\midrule

Grounding DINO detection threshold
&
$0.25$
&
Minimum confidence required for open-vocabulary candidate detections.
\\

Effective object range $r_{\rm obj}$
&
$2.0\ \mathrm m$
&
Maximum horizontal distance for completing ordinary object-approach
constraints.
\\

Object temporal support $K_{\rm obj}$
&
$5$
&
Number of semantic perception events used for temporal denoising of object
constraints.
\\

Region debounce $\kappa_{\rm in},\kappa_{\rm out}$
&
$3$
&
Number of consecutive observations required for stable location entry and
exit.
\\

Semantic translation trigger $\tau_p$
&
$0.30\ \mathrm m$
&
Maximum translation interval between routine semantic perception refreshes.
\\

Semantic rotation trigger $\tau_\psi$
&
$30^\circ$
&
Maximum heading change between routine semantic perception refreshes.
\\

Forward-motion angular boundary
&
$15^\circ$
&
Maximum deviation from the reference heading for forward displacement
classification.
\\

Backward-motion angular boundary
&
$120^\circ$
&
Minimum deviation from the reference heading for backward displacement
classification.
\\

\midrule
\multicolumn{3}{c}{\textit{Mapping and Navigation}} \\
\midrule

Map resolution
&
$0.05\ \mathrm{m/cell}$
&
Spatial resolution of the world-aligned bird's-eye-view map.
\\

Waypoint arrival threshold $d_{\rm wp}$
&
$0.25\ \mathrm m$
&
Distance below which an intermediate waypoint is considered reached.
\\

Safe-traversal margin $d_{\rm safe}$
&
$0.25\ \mathrm m$
&
Obstacle-clearance margin used when constructing the safe reachable region.
\\

Final-target distance threshold
&
$3.00\ \mathrm m$
&
Distance criterion used with final constraint completion before issuing
STOP.
\\

Cycle pose threshold $d_{\rm cyc,pos}$
&
$0.50\ \mathrm m$
&
Maximum position distance for matching a previous navigation decision state.
\\

Cycle waypoint threshold $d_{\rm cyc,wp}$
&
$0.30\ \mathrm m$
&
Maximum waypoint distance for identifying a repeated waypoint decision.
\\

Yaw-oscillation threshold $\delta_\psi$
&
$25^\circ$
&
Minimum per-step heading change counted by the oscillation detector.
\\

Maximum recovery attempts $R_{\max}$
&
$3$
&
Maximum number of recovery attempts before an episode is declared
unrecoverable.
\\

\bottomrule
\end{tabularx}
\end{table}

\end{document}